\documentclass[review]{elsarticle}

\usepackage{lineno,hyperref}
\modulolinenumbers[5]

\journal{}

\usepackage[english,activeacute]{babel}
\usepackage{amsfonts}
\usepackage{amsmath}
\usepackage{url}
\usepackage{hyperref}
\usepackage{xcolor}
\usepackage{empheq}
\usepackage{algorithm}
\usepackage{algpseudocode}
\usepackage{multirow}
\usepackage{booktabs}
\usepackage{subfig}
\usepackage{graphicx}
\usepackage{floatpag}

\newcommand{\myreferences}{Bibliography_all}
\newcommand\asunst{\bgroup\markoverwith{\revisionone{\rule[0.5ex]{2pt}{0.4pt}}}\ULon}
\newcommand{\revisionone}[1]{\textcolor{black}{#1}}
\newcommand{\mylabelcolorrevisedfiguresandtables}{black}

\begin{document}
\renewcommand{\arraystretch}{0.7}
\begin{frontmatter}

\title{A novel embedded min-max approach for feature selection in nonlinear \revisionone{s}upport \revisionone{v}ector \revisionone{m}achine classification}

\author{Asunci\'on Jim\'enez-Cordero\corref{mycorrespondingauthor}}
\cortext[mycorrespondingauthor]{Corresponding author}
\ead{asuncionjc@uma.es}
\author{Juan Miguel Morales}
\ead{juanmi82mg@gmail.com}
\author{Salvador Pineda}
\ead{spinedamorente@gmail.com}

%\address[mymainaddress]{OASYS Group. Ada Byron Research Building,\\ C/ Arquitecto Francisco Pe\~nalosa 18, 29010,\\ University of M\'alaga, M\'alaga, Spain }
 \address{OASYS Group, University of M\'alaga, M\'alaga, Spain }

\begin{abstract}
In recent years, feature selection has become a challenging problem in several machine learning fields, \revisionone{such as} classification problems. \revisionone{Support Vector Machine (SVM) is a well-known technique applied in classification tasks.} Various methodologies have been proposed in the literature to select the most relevant features in SVM. Unfortunately, all of them either deal with the feature selection problem in the linear classification setting or propose ad-hoc approaches that are difficult to implement in practice. In contrast, we propose an embedded feature selection method based on a min-max optimization problem, where a trade-off between model complexity and classification accuracy is sought. By leveraging duality theory, we equivalently reformulate the min-max problem and solve it without further ado using off-the-shelf software for nonlinear optimization. The efficiency and usefulness of our approach are tested on several benchmark data sets in terms of accuracy, number of selected features and interpretability. 
\end{abstract}


\end{frontmatter}

%\linenumbers
  
  \section{Introduction} \label{sec: Introduction}
  
In the era of big data, where huge quantities of information are collected every day, the problem of determining which of these data are really important is a challenging task. Indeed, in recent years, the number of processes where hundreds or even thousands of features are collected has considerably increased. Here, a feature is defined as an individual and measurable property of the process. Thus, it is desirable to apply machine learning techniques to retrieve knowledge from data, or equivalently to know which are the most informative features, i.e., to perform feature selection. Hence, no wonder that feature selection algorithms are on trend, \cite{bertolazzi2016integer, li2019key}.

The goal of feature selection is to remove the most irrelevant and redundant features to obtain an optimal feature subset. Feature selection has three main advantages: First, it enhances the interpretability of the results by building simpler models, \cite{blanquero2017functional,munoz2020informative}. Second, it reduces the noise and leads to cleaner and more understandable data, \cite{bolon2015recent,byeon2008simultaneously}. Finally, it may improve the prediction performance since overfitting is reduced, \cite{blanquero2017variable,lee2015kernel}. 
 
A comprehensive description of different feature selection methods, including some examples and a brief discussion on their stability, is done in \cite{chandrashekar2014survey}. For the most recent surveys on this topic, the reader is referred to \cite{li2017feature, li2017recent}. 

Feature selection techniques can be applied in both regression, \cite{andersen2010variable}, and classification algorithms, \cite{bertolazzi2016integer,tang2014feature}. In this paper, we focus on feature selection methods for the well-known Support Vector Machine (SVM) binary classification problem, \cite{cristianini2000introduction}. In plain words, SVM aims at finding the hyperplane that maximizes the minimum distance of the training points of different classes. 

Feature selection techniques are usually classified into \emph{filter}, \emph{wrapper} and \emph{embedded} methods, \cite{chandrashekar2014survey}. Filter methods act on the data without taking into account the machine learning technique that will be used to extract knowledge from them.  Consequently, they are usually applied as a preprocessing step. They rank all the features according to a score function computed from the data and filter out the lowly ranked variables. While filter methods are computationally fast and simple, they completely ignore the interaction with the learning approach.

Wrapper methods aim to find a subset of variables that gives the best predictor performance value. Two steps are performed in a wrapper method. First, a subset of features is selected, and second, the quality of such a subset is evaluated according to a score function based on the machine learning problem. This process is repeated until some stopping criterion is satisfied. Here, the learning machine acts as a black-box, but it somehow guides the final results. Nevertheless, since most of the computational time is spent on training the model, this type of feature selection method is rarely applied.

Finally, embedded methods simultaneously train the model and perform feature selection. That is, the learning part and the feature selection part are simultaneously performed.  Therefore, they can be deemed as more ``clever'' methods than the filter strategies since they interact with the prediction model, and faster than the wrapper methods since the learning model does not need to be trained every time a feature subset is selected. The feature selection method for nonlinear SVM classification that we propose falls within this category.

Several works on the topic of embedded feature selection methods for SVM are currently available in the technical literature. In the linear setting, for instance, \revisionone{we can highlight the $\ell_1$-regularization SVM model, \cite{nguyen2010optimal, zhu20041}, where the $\ell_1$-norm of the coefficients 
instead of the usual $\ell_2$ norm is minimized to get sparse solutions. The number of features is automatically tuned via a bisection method in the optimization problem in \cite{ghaddar2018high}, whose formulation takes into account that this number cannot exceed a prefixed value. Following the idea of controlling the number of selected features by way of a user-specified parameter, we highlight the work of \cite{labbe2019mixed}, where they present a mixed-integer linear program with a budget constraint on such a number. This formulation also includes} two big-M-type constraints that link the continuous and binary variables. In order to guarantee tight enough values of such big Ms, the authors propose several strategies that imply the resolution of extra optimization problems before solving the feature selection program. To our knowledge, \revisionone{a bilevel optimization problem to perform an embedded feature selection method was for the first time proposed in }\cite{kunapuli2008classification}. The Karush-Kuhn-Tucker conditions are used to reformulate it, and its relaxed version is solved with off-the-shelf nonlinear solvers. \revisionone{This work has been extended in} \cite{agor2019feature} by adding new binary variables in the upper-level problem to control the number of selected features. 

Regarding embedded models for nonlinear SVM, most of them are based on regularization approaches, which trade-off between the SVM learning objective and the complexity of the resulting classification model. For example, \revisionone{a penalization of the objective function of the dual of the SVM problem with an approximation of the 0-``norm'' of the feature vector is proposed in \cite{maldonado2011simultaneous, zhu2018embedded}}. An ad-hoc strategy where six hyperparameters should be carefully tuned is proposed as the solving strategy.  Four regularization-based approaches are formulated in \cite{neumann2005combined}. DC (difference of convex functions) techniques are proposed to solve them. Another regularization strategy is given in \cite{allen2013automatic}, where a $\ell_1$-penalty term is added to the objective function. To solve such a problem, the kernel is linearized with respect to the feature weights to obtain a biconvex problem (in the classification and feature selection variables), which is then solved with gradient techniques in an alternating algorithm. The above-mentioned articles have proposed regularization methods with continuous variables. In contrast, a regularization model with binary variables, indicating if a feature is removed or not, is built in \cite{weston2001feature}. In particular, the authors of \cite{weston2001feature} perform feature selection by introducing binary variables in a model that minimizes a radius-margin bound on the leave-one-out error of the hard-margin SVM. The $\{0, 1\}$-variables are then relaxed to solve a penalized version of the proposed optimization problem via gradient-based approaches. In such a new model, the objective function now includes a penalization which controls, via a parameter conveniently tuned, the number of retained features. They also add a constraint which fixes the number of variables to be used. As for the approaches where binary variables are utilized to determine if a feature is removed or not, we should highlight the work of  \cite{mangasarian2007feature}, where a mixed-integer nonlinear model with binary variables is built. The binary variables are iteratively updated according to a kernel-based classifier. 

To sum up, the above-mentioned approaches can be classified into two groups. The first type of methodologies deal with the feature selection problem in linear SVM, and therefore, they are unable to handle nonlinear separable data sets. The second group of references select the most relevant features in the nonlinear SVM classification at the expense of requiring ad-hoc solution algorithms with limited performance guarantees. To the best of our knowledge, none of the existing research works on this topic proposes a feature selection method for nonlinear SVM classification that can be seamlessly solved using off-the-shelf optimization software. 

\revisionone{In this paper, we propose a new embedded method for feature selection in nonlinear SVM classification that is efficiently solved using off-the-shelf optimization software. Hence, the contributions of our work are:}

\revisionone{- We formulate a min-max optimization problem whereby we balance two objectives, namely, model complexity via a norm of the feature weights and a proxy of the number of well-classified individuals expressed in terms of the SVM margin.}

\revisionone{- By way of duality theory, we reformulate this problem as a single-level equivalent optimization problem that can be efficiently processed by off-the-shelf nonlinear solvers. This way, we avoid the use of arduous ad-hoc solution strategies usually applied in the literature that, besides, often rely on the tuning of multiple hyperparameters.}

\revisionone{- We conduct a series of numerical experiments using various data sets to assess our approach and compare it with alternative state-of-the-art methods available in the technical literature. These experiments show that our approach yields similar or even better results than those alternative methods, with the added distinct advantage of being simpler and easily implementable.}

The remaining of this paper is structured as follows: Section \ref{sec: SVM Classification} briefly introduces basic definitions and concepts regarding SVM. Section \ref{sec: Feature Selection Methodology} formulates the proposed optimization problem and explain how to solve it. Section \ref{sec: Experimental Setup} is devoted to the description of the data sets, experiments and comparative algorithms, and Section \ref{sec: Numerical Experience} details the numerical experience performed. We finish in Section \ref{sec: Conclusions and Future Work} with some conclusions and possible extensions.
  
 \section{SVM Classification}\label{sec: SVM Classification}
 We focus on the binary classification problem: Given two groups of labelled data, the aim is to predict the label of an unobserved point based on the knowledge extracted from the training points. More precisely, consider a sample of individuals $\mathcal{S}$. For each individual $i\in\mathcal{S}$, we associate the pair $(x_i, y_i)$, where $x_i\in \mathbb{R}^M$ is a $M-$dimensional vector representing the features, and $y_i\in\{-1, +1\}$ denotes the label of the individual $i$. The main goal is to find a classification rule to predict the label $y$ of a new unseen individual using the information provided by $x$. 
 
Several strategies have been developed in the literature to handle the binary classification problem. See \cite{kotsiantis2006machine} for a review. In this paper, we apply the well-known and extensively used method known as \emph{Support Vector Machine} (SVM),  \cite{cristianini2000introduction}.  The primal formulation of the soft-margin SVM problem is as follows:
 \begin{subequations}\label{eq: primal SVM}
  \begin{empheq}[left=\empheqlbrace]{align}
    \min\limits_{w, b, \xi} & \,\frac{1}{2} \revisionone{\|w\|^2_2} + C\sum\limits_{i \in \mathcal{S}} \xi_i\\
    \text{s.t. }  & y_{i}(w'x_i + b) \geq 1 - \xi_i, \forall i\\
    &\xi_i\geq 0,  \forall i
    \end{empheq}
  \end{subequations}
  
 \noindent where the best separating hyperplane has the form $w'x + b = 0$. The normal vector to the hyperplane is denoted by $w\in\mathbb{R}^M$, $b$ indicates a threshold value, the prime denotes the transpose, e.g., $w'$\revisionone{, and $\|w\|_2^2$ is the squared of the $\ell_2$-norm of $w$}. In addition, the notation $r's$ indicates the dot product between the vectors $r$ and $s$, i.e., $r's=\sum\limits_i r_is_i$. Finally, a regularization parameter $C$ is introduced to penalize the misclassified points via the slack variables $\xi_i$, $\forall i$.
  
The following linear classification rule is derived from the optimal solution of Problem \eqref{eq: primal SVM}: A new unseen point $x$ is classified in class $1$ if and only if $\hat{y}(x) + b \geq 0$, where $\hat{y}(x)$ is the so-called \emph{score} function, defined as:
 \begin{equation}\label{eq: score function primal SVM}
 \hat{y}(x) = w'x
 \end{equation}
 
SVM cannot only handle linear binary classification problems but is also able to deal with nonlinear separable training points by means of the \emph{kernel trick}. The key idea is to translate the original data $x_i$ to a higher dimensional space $\mathcal{X}$ through a feature map $\phi: \mathbb{R}^M\rightarrow\mathcal{X}$, where the data become linear. Hence, Problem \eqref{eq: primal SVM} can be written in terms of the transformed data, $\phi(x_i)$ as follows:
  \begin{subequations}\label{eq: primal SVM phi}
  \begin{empheq}[left=\empheqlbrace]{align}
    \min\limits_{w, b, \xi} & \,\frac{1}{2} \revisionone{\|w\|^2_2} + C\sum\limits_{i \in \mathcal{S}} \xi_i\\
    \text{s.t. }  & y_{i}(w'\phi(x_i) + b) \geq 1 - \xi_i, \forall i\\
    &\xi_i\geq 0,  \forall i
    \end{empheq}
  \end{subequations}
 
The score function \eqref{eq: score function primal SVM} is, therefore, modified as indicated below:
 \begin{equation}\label{eq: score function primal SVM phi}
 \hat{y}(x) = w'\phi(x)
 \end{equation}
 
It is quite obvious that the nonlinear case can be reduced to the linear one, just setting $\phi(x) = x$. Unfortunately, the expression of $\phi$ is usually unknown and consequently,  Problem \eqref{eq: primal SVM phi} cannot be solved in practice. However, this issue is handled by resorting to the dual formulation of \eqref{eq: primal SVM phi}:
 \begin{subequations}\label{eq: dual SVM phi}
\begin{empheq}[left=\empheqlbrace]{align}
     \max\limits_{\alpha}& \sum\limits_{i \in \mathcal{S}} \alpha_i - \frac{1}{2}\sum\limits_{i, \ell}\alpha_i\alpha_{\ell}y_{i}y_{\ell} \phi(x_i)' \phi(x_{\ell})\\
    \text{s.t. }  & \sum\limits_{i \in \mathcal{S}} \alpha_{i}y_i = 0, \label{subeq: sum_alpha_times_label_equal_zero}\\
    & 0\leq \alpha_{i} \leq C,  \forall i  
    \end{empheq}
   \end{subequations}
   
Problem \eqref{eq: dual SVM phi} maximizes a quadratic concave objective function with linear constraints. Hence, it can be solved with standard convex optimization solvers. Moreover, as a consequence of the Lagrange dual reformulation of Problem \eqref{eq: primal SVM phi}, it holds that the coefficients of the hyperplane normal vector, $w$, can be expressed in terms of the $\alpha$ variables, as follows:
\begin{equation}\label{eq: w in terms of alpha}
w = \sum\limits_{i\in \mathcal{S}} \alpha_iy_i\phi(x_i)
\end{equation}

\noindent and, therefore, the score function $\hat{y}(x)$ in \eqref{eq: score function primal SVM phi} turns out to be:
\begin{equation}\label{eq: score function dual SVM phi}
 \hat{y}(x) =  \sum\limits_{i\in \mathcal{S}} \alpha_iy_i\phi(x_i)' \phi(x)
 \end{equation}
 
Note that both the resolution of Problem \eqref{eq: dual SVM phi} and the evaluation of the score function in \eqref{eq: score function dual SVM phi} do not depend on computing the value of $\phi$ (whose explicit form is unknown), but on computing the value of the dot product $\phi(x_i)'\phi(x_{\ell})$, $\forall (i, \ell)$. This tremendously simplify the calculation of a nonlinear classification rule by way of SVM.  Indeed, it suffices to select a so-called \emph{kernel function}, $K: \mathbb{R}^M\times\mathbb{R}^M\rightarrow \mathbb{R}$, as such a dot product, i.e.:
\begin{equation}\label{eq: kernel as dot product}
K(x_i, x_{\ell}) =\phi(x_i)' \phi(x_{\ell}), \quad \forall i, \ell
\end{equation}
 
Consequently, the score function \eqref{eq: score function dual SVM phi} can be written as:
\begin{equation}\label{eq: score function dual SVM phi kernel}
 \hat{y}(x) =  \sum\limits_{i\in \mathcal{S}} \alpha_iy_i K(x_i, x_{\ell})
 \end{equation}
 \noindent and Problem \eqref{eq: dual SVM phi} can be recast as:
 \begin{subequations}\label{eq: dual SVM}
\begin{empheq}[left=\empheqlbrace]{align}
     \max\limits_{\alpha}& \sum\limits_{i \in \mathcal{S}} \alpha_i - \frac{1}{2}\sum\limits_{i, \ell}\alpha_i\alpha_{\ell}y_{i}y_{\ell}K(x_i, x_{\ell})\\
    \text{s.t. }  & \sum\limits_{i \in \mathcal{S}} \alpha_{i}y_i = 0, \forall i \label{subeq: sum_alpha_times_label_equal_zero}\\
    & 0\leq \alpha_{i} \leq C,  \forall i  
    \end{empheq}
   \end{subequations}

Intuitively, the kernel function implicitly gives us access to a (possibly infinite dimensional) family of feature mappings $\phi(\cdot)$ without actually having to work with them. In the next section, we elaborate on how we propose to enrich problem \eqref{eq: dual SVM} with the ability to automatically perform feature selection.   
   
\section{Feature Selection Methodology}\label{sec: Feature Selection Methodology}

This section details the proposed approach to select the most relevant features when classifying. Particularly, in Section \ref{subs: Model Formulation}, we formulate a min-max optimization program to solve the feature selection problem. Section \ref{subsec: Model Reformulation} explains how to reformulate such a problem to be then solved with off-the-shelf software in Section \ref{subs: Resolution Strategy}.

  \subsection{Min-max problem formulation for feature selection} \label{subs: Model Formulation}
 The kernel trick in the dual formulation \eqref{eq: dual SVM} reveals that, for linearly separating the data in a certain feature space $\mathcal{X}$, it is \emph{not} necessary to know the explicit expression of the mapping $\phi(\cdot)$, but of the associated dot product or kernel $K(x_i, x_{\ell}) =\phi(x_i)' \phi(x_{\ell}), \quad \forall i, \ell$. Actually, it can be considered that all the maps $\phi(\cdot)$ reproduced by the same kernel are equivalent, \cite{mercer2006minh}. No wonder, therefore, that the success of the nonlinear SVM problem highly relies on a careful selection of the kernel. In this vein, our approach aims to identify a kernel whereby the SVM is able to separate the data (as much as possible) using the most informative features only.  
  
  Following this argument, consider next  a family of kernels $K_{\gamma}(\cdot, \cdot)$ parameterized in terms of a finite-dimensional vector $\gamma$. Each member in this family reproduces a catalog of feature maps $\phi(\cdot)$ in a feature space $\mathcal{F}_{\gamma}$. We can now reformulate the primal version of the nonlinear SVM to account for this additional degree of freedom as follows:
    \begin{subequations}\label{eq: primal SVM phi min phi}
  \begin{empheq}[left=\empheqlbrace]{align}
    \min\limits_{\gamma}\ \min\limits_{w, b, \xi} & \,\frac{1}{2} \revisionone{\|w\|^2_2} + C\sum\limits_{i \in \mathcal{S}} \xi_i\label{eq: primal SVM phi min phi_OF}\\
    \text{s.t. }  & y_{i}(w'\phi(x_i) + b) \geq 1 - \xi_i, \forall i\\
    &\xi_i\geq 0,  \forall i,\\
    & \phi \in \mathcal{F}_{\gamma},
    \end{empheq}
  \end{subequations}
 
 \noindent where we also look for the best $\gamma$, i.e., the best functional space $\mathcal{F}_{\gamma}$, that maximizes the SVM soft margin.

 Unsurprisingly, problem \eqref{eq: primal SVM phi min phi} is intractable, now not only because the particular form of the feature map $\phi(\cdot)$ is unknown, but also because we do not have an explicit expression of the feature space $\mathcal{F}_{\gamma}$ in terms of the parameter vector $\gamma$. In a first step to cope with this challenge, we resort again to the dual formulation of the nonlinear SVM, that is:
 \begin{subequations}\label{eq: min gamma dual SVM}
\begin{empheq}[left=\empheqlbrace]{align}
     \min\limits_{\gamma}\max\limits_{\alpha}& \sum\limits_{i \in \mathcal{S}} \alpha_i - \frac{1}{2}\sum\limits_{i, \ell}\alpha_i\alpha_{\ell}y_{i}y_{\ell}K_{\gamma}(x_i, x_{\ell})\label{eq: min gamma dual SVM_OF}\\
    \text{s.t. }  & \sum\limits_{i \in \mathcal{S}} \alpha_{i}y_i = 0, \forall i \label{subeq: sum_alpha_times_label_equal_zero}\\
    & 0\leq \alpha_{i} \leq C,  \forall i
    \end{empheq}
   \end{subequations}
   
 As opposed to \eqref{eq: primal SVM phi min phi}, the good thing about problem \eqref{eq: min gamma dual SVM} is that we may indeed have an explicit expression of a kernel $K_{\gamma}(\cdot,\cdot)$ in terms of a finite-dimensional parameter vector $\gamma$. For instance, in this paper, we will work with the anisotropic Gaussian kernel with bandwidth parameter $\gamma \geq 0$, which is well known for its flexibility and which takes the following form: 
   \begin{equation} \label{eq: kernel expression}
      K_{\gamma}(x_i, x_{\ell}) =\exp\left(-\sum\limits_{j = 1}^M \gamma_j (x_{ij} - x_{\ell j})^2\right)
  \end{equation}
 
In particular, the anisotropic Gaussian kernel will allow us to perform feature selection via feature weighting in a natural way. In effect, using the expression in \eqref{eq: kernel expression}, the importance of each feature can be easily measured through the value of $\gamma_j$. More precisely, values of $\gamma_j$ tending to zero imply that the associated feature $j$ plays no role in the classification. In contrast, larger values of $\gamma_j$ indicate that feature $j$ is critical for obtaining good classification results. 

A major drawback of problem \eqref{eq: min gamma dual SVM} (or equivalently, of problem \eqref{eq: primal SVM phi min phi}) is, however, that the additional degree of freedom introduced by the parameter vector $\gamma$ is likely to produce overfitting of the training data. Indeed, it is known that values of $\gamma_j$ tending to infinite in the Gaussian kernel \eqref{eq: kernel expression} will lead to this troublesome phenomenon. Consequently, we somehow need to penalize large values of $\gamma$ in problem \eqref{eq: min gamma dual SVM}. At the same time, if $\gamma_j = 0, \forall j$, then the kernel expression \eqref{eq: kernel expression} is equal to one for all pair of individuals in the sample, i.e., $K_{\gamma} (x_i, x_{\ell}) = 1, \forall i, \ell$. It is thus easy to check by combining the SVM classification rule \eqref{eq: score function primal SVM}, the score function \eqref{eq: score function dual SVM phi kernel} and constraint \eqref{subeq: sum_alpha_times_label_equal_zero} that the predicted label of a new unseen individual, in this case, will always coincide with the sign of $b$, resulting in poor classification performance. 

Hence, it is quite apparent that there exists a trade-off between model complexity and classification accuracy. In other words, a model that simultaneously minimizes the feature weights and maximizes the accuracy is desired. To this aim, we modify Problem \eqref{eq: min gamma dual SVM} to propose the min-max optimization problem \eqref{eq: fs non linear SVM}, where a trade-off between two objectives, namely the $p$-(pseudo)norm of the feature weights vector $\gamma$, $\|\gamma\|^p_p$, for $p\geq 0$, and the objective function of the SVM problem \eqref{eq: dual SVM} is to be optimized. The importance associated with each objective is measured in terms of parameter $C_2$, which balances the complexity of the model and the classification accuracy and whose value depends on the user's preferences. Values of $C_2$ close to $0$ favor models with high in-sample prediction accuracy even though the number of features to be used is large. Conversely, $C_2$ values tending to $1$ result in models with a reduced number of features at the expense of sacrificing some accuracy.
  \begin{subequations}\label{eq: fs non linear SVM}
 \begin{empheq}[left=\empheqlbrace]{align}
   \min\limits_{\gamma\geq0} & \,\left[ C_2 \|\gamma\|_p^p  + (1-C_2)\max\limits_{\alpha} \sum\limits_{i\in\mathcal{S}} \alpha_i - \frac{1}{2}\sum\limits_{i, \ell \in \mathcal{S}}\alpha_i\alpha_{\ell}y_{i}y_{\ell}K_{\gamma}(x_i, x_{\ell}) \right]\\
   &\qquad \qquad \qquad \qquad \; \; \; \; \; \text{s.t. }   \sum\limits_{i \in \mathcal{S}} \alpha_{i}y_i = 0\\
   & \qquad \qquad \qquad \qquad \qquad \quad \; \;  0\leq \alpha_{i} \leq C,  \forall i
  \end{empheq}
  \end{subequations}
  
 Compared to the rest of embedded approaches for nonlinear SVM classification in the technical literature, our model \eqref{eq: fs non linear SVM} does not select the most relevant features via binary variables, as in \cite{mangasarian2007feature, weston2001feature}. In contrast, we perform feature selection through feature weighting by means of continuous variables. This way, we do not only avoid the difficulties associated to Integer Programming, but we also get to know whether a feature is relevant or not, together with some measure of its degree of importance.
 
 Furthermore, all the models proposed in \cite{allen2013automatic,maldonado2011simultaneous, neumann2005combined} penalizes the size of the feature vector \emph{within} the objective function of the dual SVM problem \eqref{eq: dual SVM}. However, our goal is to find the $\gamma$ that leads to the largest SVM margin. This implies minimizing over $\gamma$ the minimum of $\frac{1}{2} \|w\|^2 + C\sum\limits_{i \in \mathcal{S}} \xi_i$ over $w$ and $\xi$, as in~\eqref{eq: primal SVM phi min phi}, or equivalently, minimizing over $\gamma$ the maximum of $\sum\limits_{i \in \mathcal{S}} \alpha_i - \frac{1}{2}\sum\limits_{i, \ell}\alpha_i\alpha_{\ell}y_{i}y_{\ell}K_{\gamma}(x_i, x_{\ell})$ over $\alpha$, as in \eqref{eq: min gamma dual SVM}. Since we still want to leverage the kernel trick, we need to opt for the latter and introduce the regularization term accordingly as in \eqref{eq: fs non linear SVM}. This gives rise to the min-max approach for feature selection that we propose. 
 
\subsection{Model Reformulation}\label{subsec: Model Reformulation}
Problem \eqref{eq: fs non linear SVM} is a nonconvex optimization problem very hard to solve, in general. The aim of this section is to reformulate such a problem in order to solve it via off-the-shelf software.
 
First of all, we equivalently rewrite Problem \eqref{eq: fs non linear SVM} using its epigraph form as:
\begin{subequations}\label{eq:bilevel non linear SVM}
 \begin{empheq}[left=\empheqlbrace]{align}
   \min\limits_{\gamma\geq0,\, z} &\; C_2\|\gamma\|_p^p +(1-C_2) z&&\\
   \text{s.t. }  & \; z\geq  \max\limits_{\alpha} \sum\limits_{i \in \mathcal{S}} \alpha_i - \frac{1}{2}\sum\limits_{i, \ell \in \mathcal{S}}\alpha_i\alpha_{\ell}y_{i}y_{\ell}K_{\gamma}(x_i, x_{\ell}) \label{subeq: objective function lower level bilevel non linear SVM}\\ 
  &\qquad \; \text{s.t. }   \sum\limits_{i\in\mathcal{S}} \alpha_{i}y_i = 0 & (\nu) \label{subeq: dual_equality_constraint_SVM}\\
   &\qquad \qquad \;   0\leq \alpha_{i} \leq C, \,  \forall i  & (\lambda^0_i, \lambda^C_i) \label{subeq: dual_inequality_constraints_SVM}&& 
  \end{empheq}
  \end{subequations}

Problem \eqref{eq:bilevel non linear SVM} can be seen as a bilevel optimization problem where the upper-level problem aims at obtaining good classification results with a low number of features, whereas the lower-level problem focuses on the classification task. Indeed, the lower-level problem states that the decision variable $z$ is lower-bounded by the optimal solution of the dual SVM problem \eqref{eq: dual SVM}.

In order to solve Problem \eqref{eq:bilevel non linear SVM}, we propose a reformulation based on the lower-level dual problem, which exploits the fact that the SVM problem \eqref{eq: dual SVM} is a convex optimization problem with a quadratic objective function and affine constraints. Hence, strong duality holds and the lower-level problem \eqref{subeq: objective function lower level bilevel non linear SVM} - \eqref{subeq: dual_inequality_constraints_SVM} can be equivalently replaced by its dual, \cite{convex2004boyd}. Actually, strong duality also allows us to justify the outer minimization in \eqref{eq: fs non linear SVM} using arguments from Mathematical Programming: Maximizing the SVM soft margin involves minimizing objective function \eqref{eq: primal SVM phi min phi_OF}, which, in turn, takes on the same value as that of the dual objective \eqref{eq: min gamma dual SVM_OF}  at the primal and dual optima. Therefore, we are to minimize the dual objective \eqref{eq: min gamma dual SVM_OF} over $\gamma$.

 We start then building the Lagrangian function of the lower-level problem. For the sake of simplicity, in what follows, matrix notation will be used. We define $G_{\gamma} := diag(y)K_{\gamma}diag(y)$ as the quadratic form of the SVM problem, and $diag(y)$ represents the matrix with the vector $y$ in its diagonal. Moreover, $e$ represents a vector full of ones of appropriate dimension, and the variables between brackets next to the constraints \eqref{subeq: dual_equality_constraint_SVM} and \eqref{subeq: dual_inequality_constraints_SVM} are their corresponding dual decision variables. With this notation, the Lagragian function of the lower-level problem is computed as follows:
\begin{equation}\label{eq: lagrangian dual lower level}
\mathcal{L}(\alpha, \nu, \lambda^0, \lambda^C) = e'\alpha -\frac{1}{2}\alpha'G_{\gamma}\alpha - \nu y'\alpha+(\lambda^0)'\alpha - (\lambda^C)'(\alpha - Ce)
\end{equation}

To compute the objective function of the dual of the lower-level problem, it is necessary to compute the gradient of $\mathcal{L}(\alpha, \nu, \lambda^0, \lambda^C)$ with respect to $\alpha$, $\nabla_\alpha\mathcal{L}(\alpha, \nu, \lambda^0, \lambda^C)$ and let it vanish, i.e.:
\begin{equation}\label{eq: lagrangian gradient dual lower level}
\nabla_\alpha\mathcal{L}(\alpha, \nu, \lambda^0, \lambda^C) =  e -G_{\gamma}\alpha - \nu y+\lambda^0 - \lambda^C = 0
\end{equation}

Therefore, the dual formulation of the SVM problem \eqref{eq: dual SVM} is:
 \begin{subequations}\label{eq: dual lower level bilevel gradient equal zero}
  \begin{empheq}[left=\empheqlbrace]{align}
   \min\limits_{\alpha, \nu, \lambda^0, \lambda^C} & \, -\frac{1}{2}\alpha'G_{\gamma}\alpha +(e-\nu y + \lambda^0-\lambda^C)'\alpha + C(\lambda^C)'e\\
   \text{s.t. } & \, G_{\gamma}\alpha -(e-\nu y + \lambda^0-\lambda^C) = 0\\
   & \, \lambda^0, \lambda^C \geq0
   \end{empheq}
  \end{subequations}
  
\noindent and Problem \eqref{eq:bilevel non linear SVM} is equivalent to:
 \begin{subequations}\label{eq:bilevel non linear SVM dual lower level}
 \begin{empheq}[left=\empheqlbrace]{align}
   \min\limits_{z, \gamma\geq0} &\; C_2\|\gamma\|_p^p +(1-C_2) z&& \label{subeq: objective function upper level bilevel non linear SVM dual lower level}\\
   \text{s.t. }  & \; z\geq  \min\limits_{\alpha, \nu, \lambda^0, \lambda^C} \, -\frac{1}{2}\alpha'G_{\gamma}\alpha +(e-\nu y + \lambda^0-\lambda^C)'\alpha + C(\lambda^C)'e\label{subeq: objective function lower level bilevel non linear SVM dual lower level}\\ 
  &\qquad \; \text{s.t. }  \, G_{\gamma}\alpha -(e-\nu y + \lambda^0-\lambda^C) = 0 \label{subeq: dual_equality_constraint_SVM dual lower level}\\
   &\qquad \qquad \;  \lambda^0, \lambda^C \geq0 \label{subeq: non negativity dual variables bilevel non linear SVM dual lower level}
  \end{empheq}
  \end{subequations}

The second term of the objective function  \eqref{subeq: objective function upper level bilevel non linear SVM dual lower level} aims at minimizing $z$, which is a variable lower-bounded by the optimal objective value of Problem \eqref{eq: dual lower level bilevel gradient equal zero}. Hence, the optimal decision variable $z$ can be replaced by the optimal value of Problem \eqref{eq: dual lower level bilevel gradient equal zero}, and Problem \eqref{eq:bilevel non linear SVM dual lower level} is written as:
 \begin{subequations}\label{eq: single level equivalent gradient equal zero}
 \small
  \begin{empheq}[left=\empheqlbrace]{align}
  \textstyle{\min\limits_{\gamma, \,\alpha, \nu, \lambda^0, \lambda^C}} & \textstyle	 C_2\|\gamma\|_p^p - (1-C_2)\left(\frac{1}{2}\alpha'G_{\gamma}\alpha -(e-\nu y + \lambda^0-\lambda^C)'\alpha - C(\lambda^C)'e\right) \label{subeq: objective function sle}\\
  \text{s.t. }  & G_{\gamma}\alpha -(e-\nu y + \lambda^0-\lambda^C) = 0 \label{subeq: equality}\\
   & \gamma, \lambda^0, \lambda^C\geq 0 \label{subeq: non negativeness}\\
   & 0\leq\alpha\leq C \label{subeq: bounds alpha}
   \end{empheq}
  \end{subequations}
  
  Problem \eqref{subeq: objective function sle}-\eqref{subeq: non negativeness} is the single-level equivalent reformulation of the bilevel optimization problem \eqref{eq:bilevel non linear SVM}. It is strongly non-convex and, as a result, we can only aspire to get local optimal solutions if nonlinear optimization solvers are used. In this regard, our numerical experiments reveal that including constraints~\eqref{subeq: bounds alpha}, albeit redundant, helps the nonlinear solver to reach a good local optimal solution faster, especially for large values of $C_2$. In Section \ref{subs: Resolution Strategy}, we elaborate on a simple but effective solving strategy based on off-the-shelf optimization software.
  
\subsection{Solving Strategy}\label{subs: Resolution Strategy}
The aim of this section is to detail the strategy carried out to solve Problem \eqref{eq: single level equivalent gradient equal zero}. We propose an efficient generic method based on grid search approaches and standard off-the-shelf solvers.
 
Firstly, we must clarify that, in order to avoid overfitting and to obtain stable results, the whole sample of individuals $\mathcal{S}$ is divided into a training and test subsamples denoted by $\tilde{\mathcal{S}}$ and $\mathcal{S}_{test}$, respectively. This process is repeated $k$ times so that there is no common individual between the test samples of two different iterations. Secondly, the proposed feature selection approach is solved for a fixed value of the hyperparameter $C_2$ defined by the user.

The first step when solving Problem \eqref{eq: single level equivalent gradient equal zero} is to determine the value of hyperparameter $C$ and to find an appropriate starting point of the $\gamma$ variable vector, $\gamma^{ini}$, for the nonlinear off-the-shelf solver. In this paper, we opt to choose the best $\gamma$ provided by the standard SVM problem \eqref{eq: dual SVM}, where no feature selection is performed, i.e., for the $\gamma$ value which gives the best predictions, when assuming that all the features play the same role and, consequently, setting $\gamma_j = \gamma, \forall j$ in the kernel function \eqref{eq: kernel expression}. To this aim, $N$-fold cross-validation has been implemented. At each iteration, sample $\tilde{\mathcal{S}}$ is divided into training and validation data, denoted respectively as $\mathcal{S}_{tr}$ and $\mathcal{S}_{val}$. Hence, for a fixed $(C, \gamma)$ varying in a grid previously selected, Problem \eqref{eq: dual SVM} is solved in $\mathcal{S}_{tr}$ with $\gamma_j = \gamma$, $\forall j$. For fixed values of $\gamma$, optimization problem \eqref{eq: dual SVM} is convex and can be solved using commercial optimization software. The selected pair $(C^*, \gamma^{ini})$ is chosen to be the one that maximizes the averaged accuracy on $\mathcal{S}_{val}$ over the $N$ folds.

Once the optimal value of $C$ and the initial solution $\gamma^{ini}$ are determined, the dual formulation of the lower-level problem \eqref{eq: dual lower level bilevel gradient equal zero} is solved in $\tilde{\mathcal{S}}$ to obtain the initial decision variables $\alpha^{ini}$, $\nu^{ini}$, $\lambda^{0, ini}$, and $\lambda^{C, ini}$. For fixed $C$ and $\gamma$, this problem is also convex and can be solved with commercial optimization software.

Next, we solve Problem \eqref{eq: single level equivalent gradient equal zero} in sample $\tilde{\mathcal{S}}$ for the same value $C^*$ and using the initial decision variables as a starting point. To do this, we use an off-the-shelf nonlinear solver. To guarantee that we work with $\alpha$ decision variables that are globally optimal for the so obtained $\gamma_j$ for all feature $j$, we then solve the convex  Problem \eqref{eq: dual SVM} in $\tilde{\mathcal{S}}$ for such a $\gamma$ vector. 

Finally, the efficiency of our approach is measured by computing the accuracy on sample $\mathcal{S}_{test}$ using the corresponding decision variables, $\alpha$ and $\gamma$, previously determined, by solving Problems \eqref{eq: dual SVM} and \eqref{eq: single level equivalent gradient equal zero}, respectively.

A pseudocode of our solving strategy for a certain division of sample $\mathcal{S}$ is sketched in Algorithm \ref{alg: solving strategy min max approach}.

\begin{algorithm}[!htb]
\small
\begin{algorithmic}[0]
\State \textbf{Input:} sample division $\tilde{\mathcal{S}}$ and $\mathcal{S}_{test}$, and hyperparameter $C_2$.
\State \textbf{\emph{Computation of initial solution}}
\For{fold in $1, \ldots, N$}
\State{$\bullet$} Define samples $\mathcal{S}_{tr}$ and $\mathcal{S}_{val}$.
\For {$(C,\gamma)$ in the grid}
\State {$\bullet$ Solve SVM problem \eqref{eq: dual SVM} on $\mathcal{S}_{tr}$ with $\gamma_j = \gamma, \, \forall j$ in kernel \eqref{eq: kernel expression}.}
\State $\bullet$ Compute accuracy on $\mathcal{S}_{val}$.
\EndFor
\EndFor
\State {$\bullet$ $(C^*,\gamma^{ini}) = \arg \max\limits_{(C, \gamma)} \text{averaged accuracy}_{\mathcal{S}_{val}}(C, \gamma)$}
\State {$\bullet$ Solve Problem \eqref{eq: dual lower level bilevel gradient equal zero}} in $\tilde{\mathcal{S}}$ for $C^*$ and starting at $\gamma^{ini}$. Obtain $\alpha^{ini}$, $\nu^{ini}$,
\State{\hspace{0.2cm}{ $\lambda^{0, ini}$, and $\lambda^{C, ini}$.}}
\State \textbf{\emph{Computation of local optimal solution}}
\State {$\bullet$ Solve Problem \eqref{eq: single level equivalent gradient equal zero} on $\tilde{\mathcal{S}}$ for $C^*$ and starting at $(\gamma^{ini}, \alpha^{ini}, \nu^{ini}, \lambda^{0, ini},\lambda^{C, ini} )$.}
\State \textbf{\emph{Computation of global optimal solution }}$\boldsymbol{\alpha}$
\State {$\bullet$ Solve Problem \eqref{eq: dual SVM} in $\tilde{\mathcal{S}}$ for fixed $\gamma$ obtained from the previous step.}
 \State \textbf{Output:} Optimal hyperparameter $
C^*$, optimal decision variables $\gamma$, $\alpha$, $\nu$,
\State \hspace*{0.01cm} $\lambda^0$ and $\lambda^C$,  and the classification accuracy on $\mathcal{S}_{test}$.
\end{algorithmic}
\caption{Solving strategy for the proposed min-max approach}\label{alg: solving strategy min max approach}
\end{algorithm}
In summary, the proposed solution strategy requires solving $N$ convex optimization problems, for each pair $(C, \gamma)$, one convex optimization problem for fixed $C^*$, and one nonconvex optimization problem. Laborious and complex ad-hoc methodologies are not necessary here.

  \section{Experimental Setup}\label{sec: Experimental Setup}
All the computational experiments carried out in this research are detailed in this section. Section \ref{subs: Data Sets Description} is devoted to the description of the data sets employed in our analyses. Section \ref{subs: Description of the Experiments} explains the experiments performed. Finally, Section \ref{subs: Comparative Algorithms} introduces the algorithms our approach is compared with.
  \subsection{Data Sets}\label{subs: Data Sets Description}
   We have worked with four databases, namely \emph{breast}, \emph{diabetes}, \emph{lymphoma}, and  \emph{colorectal}, all of which can be downloaded from \cite{OASYS2020medical}. Table \ref{table: data description} includes the number of individuals, the number of features, and the percentage of individuals of the predominant class in each of these databases. 
  
  \begin{table}[!htb]
 \centering
 \begin{tabular}{lccc}
 \toprule
 Data set &$\#$ individuals &  $\#$ features & $\%$ predominant class \\
 \midrule
\text{breast}&569 &30 & 63\% \\
\text{diabetes}&768 &8 & 65\% \\
\text{lymphoma}&96&4026&64\%\\
\text{colorectal}&62 &2000 & 65\% \\

 \bottomrule
 \end{tabular}
 \caption{Data description summary}\label{table: data description}
\end{table}

\emph{Colorectal} is known to contain outliers, \cite{bolon2014review}. Indeed, improvements in classification accuracy of up to 8-9 percentage points have been reported if those outliers are removed.  A discussion on the impact of the outliers on our approach in comparison with other strategies is detailed in Section~\ref{sec: Numerical Experience}. To this end, we have considered as outliers the 11 individuals (out of 62) identified in \cite{kadota2003detecting}. 

  \subsection{Description of the Experiments}\label{subs: Description of the Experiments}
  This section elaborates on the experiments carried out to assess and benchmark our approach, which will be denoted in Section \ref{sec: Numerical Experience} as \emph{MM-FS}.

  As a preprocessing step, the features of each data set have been normalized so that each feature belongs to the interval $[-1, 1]$. Algorithm \ref{alg: solving strategy min max approach} has been run to show the efficiency and usefulness of the proposed methodology. Indeed, in order to get stable results, the experiment given in Algorithm \ref{alg: solving strategy min max approach} has been performed $k = 10$ times. More specifically, the whole sample $\mathcal{S}$ has been divided into $10$ folds. At each iteration, $1$ out of the $10$ folds is used as the test sample $\mathcal{S}_{test}$. The remaining $9$ folds form the $\tilde{\mathcal{S}}$ sample. Note that there is no common individual between the test samples of two different iterations.
  
When computing the initial solution in Algorithm \ref{alg: solving strategy min max approach}, the $9$ folds are further subdivided into $N = 5$ folds, so that $\mathcal{S}_{tr}$ and $\mathcal{S}_{val}$ comprise, respectively, $\frac{4}{5}$ and $\frac{1}{5}$ of the data in such $9$ folds. Problem \eqref{eq: dual SVM} is solved on $\mathcal{S}_{tr}$ as indicated in Algorithm \ref{alg: solving strategy min max approach}. This process is repeated following a $5$-fold cross-validation process.

The $\gamma$ grid of the initial solution is $\{10^{-4}, \ldots, 10^4\}$ and the $p$ value chosen in the first term of the objective function \eqref{subeq: objective function sle} is $p = 1$.

As explained in Section \ref{subs: Model Formulation}, here we assume that the hyperparameter $C_2$ should be chosen by the user. Hence, in this paper, we do not provide the results of our approach for a single value of $C_2$, but a curve of the out-of-sample accuracy versus the number of features retained for different values of $C_2$ in a range. If a value of $C_2$ were to be chosen based on a specific user's criterion, then it would be selected, as usually done in the literature, according to the best results obtained in a validation sample in terms of that particular criterion.

\revisionone{Morever, we show the performance of our approach when selecting features. To this aim, for a fixed $C_2$, we provide the out-of-sample accuracy versus the number of features selected for different $C$ values.}

\revisionone{In addition, we test how consistent our approach is in the selection of features when the hyperparameter  $C_2$ is changed. For that purpose, we proceed as follows: for each value of $C_2$ and for each fold, we do a list in which each feature $j$ is ranked according to the $j$-th component, $\gamma_j$, of $\gamma$ given by Algorithm~\ref{alg: solving strategy min max approach}. Then, we check to which extent this ranking, averaged over all folds, remains unaltered for different values of $C_2$.}

\revisionone{We have also analyzed how multicollinearity affects the performance of the proposed feature selection method by checking whether the selected features are highly correlated or not.}

\revisionone{Finally}, in order to test the ability of our approach to producing interpretable SVM classification models, we also provide the name of those features which our method identifies as the most important ones and compare them with the features that are deemed as the most meaningful in the technical literature.

To sum up, our experiments provide: i) the efficient frontier ``out-of-sample classification accuracy vs norm of features weights'' that our approach is able to deliver by varying $C_2$ in the grid $\{0.01, 0.1, 0.2, \ldots, 0.8,0.9, 0.99\}$; ii) the percentage classification accuracy of our approach  \revisionone{versus the number of selected features, iii) a study of the robustness of our approach when changing hyperparameter $C_2$ with respect to the ranking of features; iv) an analysis of the selected features depending on their correlation values }and \revisionone{v}) a discussion on the level of interpretability of the selected features.

All the experiments are carried out on a cluster with $21$ Tb of RAM memory, running Suse Leap 42 Linux distribution. Models are coded in \texttt{Python 3.7} and \texttt{Pyomo 5.2} and solved using \texttt{Cplex 12.6.3} for the convex problems and \texttt{Ipopt 3.12.8} for nonconvex problems within a time limit of $24$ hours.

  \subsection{Comparative Algorithms}\label{subs: Comparative Algorithms}
  
\revisionone{Four} alternative approaches have been used to compare our proposal. The first one, denoted as \emph{NO-FS}, corresponds to the solution provided when no feature selection is made, i.e. when the importance of the features is given by a unique value $\gamma_j = \gamma,\, \forall j$ in \eqref{eq: kernel expression}. In other words, the \emph{NO-FS} method boils down to the classification results given by the initial solution of Algorithm \ref{alg: solving strategy min max approach}.

\revisionone{The second approach, named \emph{$\ell_1$-SVM}, corresponds to the SVM primal formulation where the $\ell_1$-norm is used in the regularization term instead of the standard $\ell_2$-norm. In other words, in  \emph{$\ell_1$-SVM}, Problem (1) is solved using $\|w\|_1$ instead of $\|w\|_2^2$ with \texttt{Cplex 12.6.3}.}
  
The \revisionone{third} approach, named \emph{KP-FS}, is proposed in \cite{maldonado2011simultaneous}. This model is based on a regularization of the dual SVM problem, where an approximation of the $0$-``norm'' is added as a penalty term in the objective function.
This problem is solved using a heuristic alternating algorithm that requires the careful tuning of several hyperparameters.

The results provided by \emph{KP-FS} are based on \emph{ad-hoc} solution strategies. For this reason, the feature selection approach given by reference \cite{maldonado2011simultaneous} is denoted as \emph{KP-FS ad-hoc}. Moreover, for the sake of comparison, we have also run the feature selection model \emph{KP-FS} using \emph{off-the-shelf} solvers. This comparative strategy is named \emph{KP-FS off-the-shelf}.
The values of some setting parameters used in the \emph{KP-FS off-the-shelf} methodology should be defined. For instance, the $\beta$ parameter which appears in the $0$-``norm'' approximation is set to five as the authors suggested. Both steps of the alternating approach have been solved with off-the-shelf software. In fact, the convex optimization problem of the first step is solved using \texttt{Cplex 12.6.3}, whereas \texttt{Ipopt 3.12.8} has been run to solve the nonlinear optimization problem of the second step. The maximum number of iterations of the alternating approach is set to five. In order to avoid getting stuck at local optima, a multistart with three runs is performed in the second step. The regularization parameter $C$ takes values in the set $\{10^{-4}, \ldots, 10^{-1}, 1, 2, \ldots, 9, 10, \ldots, 10^4\}$ and the hyperparameter $C_2$ ranges in the set $\{0.01, 0.1, 0.2, \ldots, 0.8,0.9, 0.99\}$.

The \revisionone{fourth} approach, denoted by \emph{MILP-FS}, has been proposed in \cite{labbe2019mixed}. The authors of \cite{labbe2019mixed} tackle the feature selection problem in linear SVM using a Mixed Integer Linear Problem (MILP), where the maximum number of selected features is chosen in advance. They devised two strategies to solve the resulting MILP, namely, a heuristic approach and an exact procedure. In both cases, extra optimization problems are to be solved.

As occurred in the \emph{KP-FS} approach, the solving strategy proposed in the \emph{MILP-FS} method is based on \emph{ad-hoc} procedures. For this reason, this method will be denoted as \emph{MILP-FS ad-hoc}. We also propose to solve the \emph{MILP-FS} model using \emph{off-the-shelf} optimization solvers. This comparative algorithm is named as \emph{MILP-FS off-the-shelf}. In this approach, the SVM parameter $C$ also moves in the set $\{10^{-4}, \ldots, 10^{-1}, 1, 2, \ldots, 9, 10, \ldots, 10^4\}$, even though our experiments show that the $C$ parameter seems not to play an important role when solving the MILP model. The budget parameter $B$, which controls the maximum number of features to be selected is set to $10, 50, 5$, and $50$ for the databases \emph{breast}, \emph{colorectal}, \emph{diabetes} and \emph{lymphoma}, in that order. The lower and upper bounds used for the big~-M constraints are fixed to $-1$ and $1$ in the data sets \emph{colorectal}, \emph{diabetes} and \emph{lymphoma}, and $-200$ and $200$ in the database \emph{breast}. This MILP model is solved using \texttt{Cplex 12.6.3} with the default options.

The division of the whole data set into samples $\tilde{\mathcal{S}}$ and $\mathcal{S}_{test}$ used in \revisionone{\emph{$\ell_1$-SVM}, }\emph{KP-FS off-the-shelf} and \emph{MILP-FS off-the-shelf} has been performed in the very same manner as done for our \emph{MM-FS} approach. Unfortunately, we do not know the exact sample division that was used for the \emph{KP-FS ad-hoc} and \emph{MILP-FS ad-hoc} approaches in \cite{maldonado2011simultaneous} and \cite{labbe2019mixed}, respectively. Thus, the results provided in Section \ref{sec: Numerical Experience} for these strategies are a \emph{verbatim} transcript of those given in \cite{maldonado2011simultaneous} and \cite{labbe2019mixed}.

\section{Numerical Experience} \label{sec: Numerical Experience}
Next, we elaborate on the results delivered by the numerical experiments we have conducted. More precisely, in Section \ref{subs: Accuracy results}, we compare the results obtained with our method in terms of classification accuracy with respect to the other algorithms previously described. Section \ref{subs: Feature selection results} discusses prediction results \revisionone{in terms of the number of selected features. Section \ref{subs: Ranking of features and multicollinearity results} studies the ranking of features derived from our approach for different $C_2$ values, as well as the effect of multicollinearity on the proposed method.} Finally, Section \ref{subs: Interpretability of the results} focuses on the interpretability of the selected features.

  \subsection{Accuracy results}\label{subs: Accuracy results}
\begin{figure}[!htb]
\thisfloatpagestyle{empty}
\centering
\subfloat[breast]{
  \includegraphics[scale = 0.3]{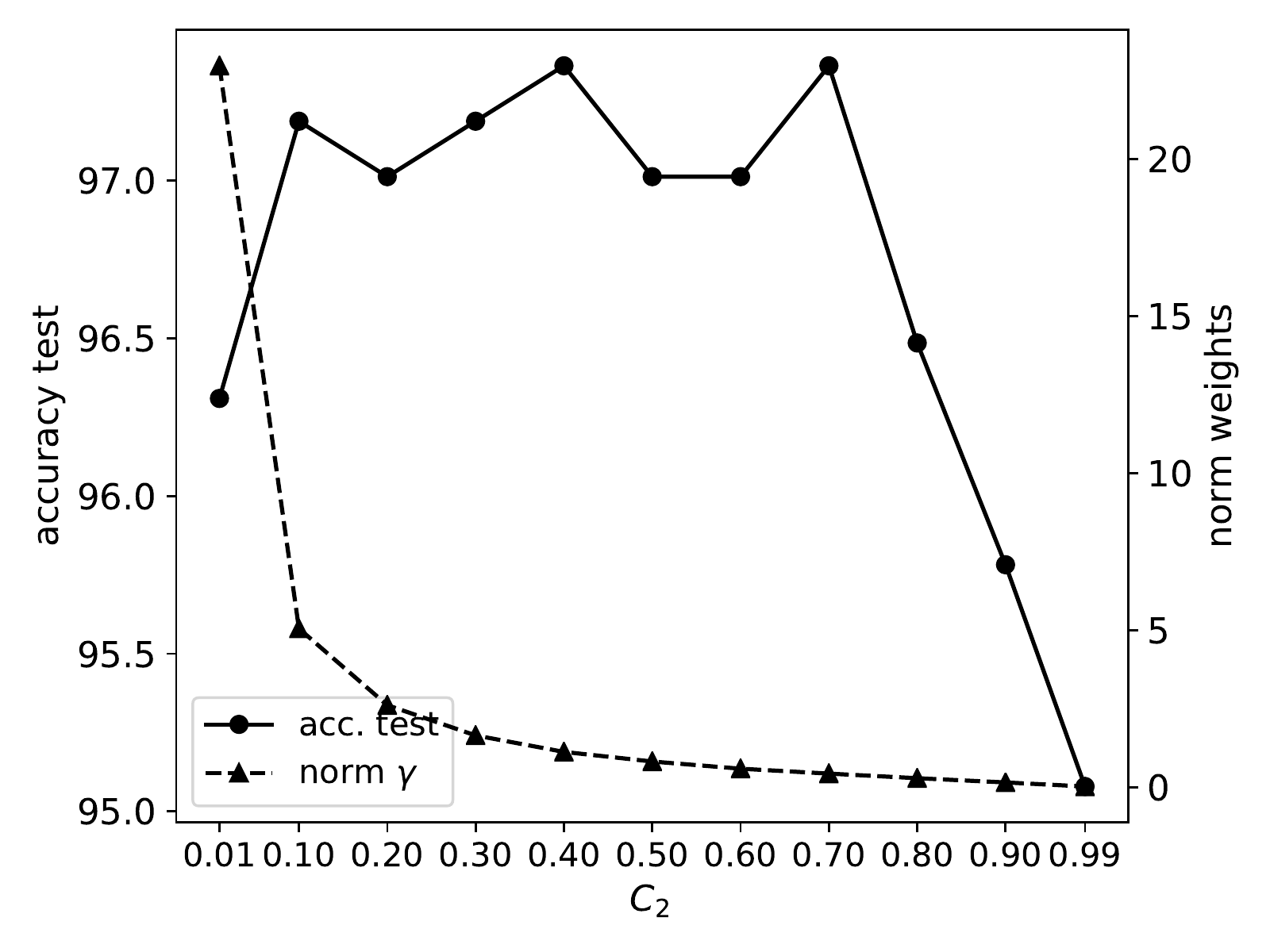}
  \label{}
}
\subfloat[diabetes]{
  \includegraphics[scale = 0.3]{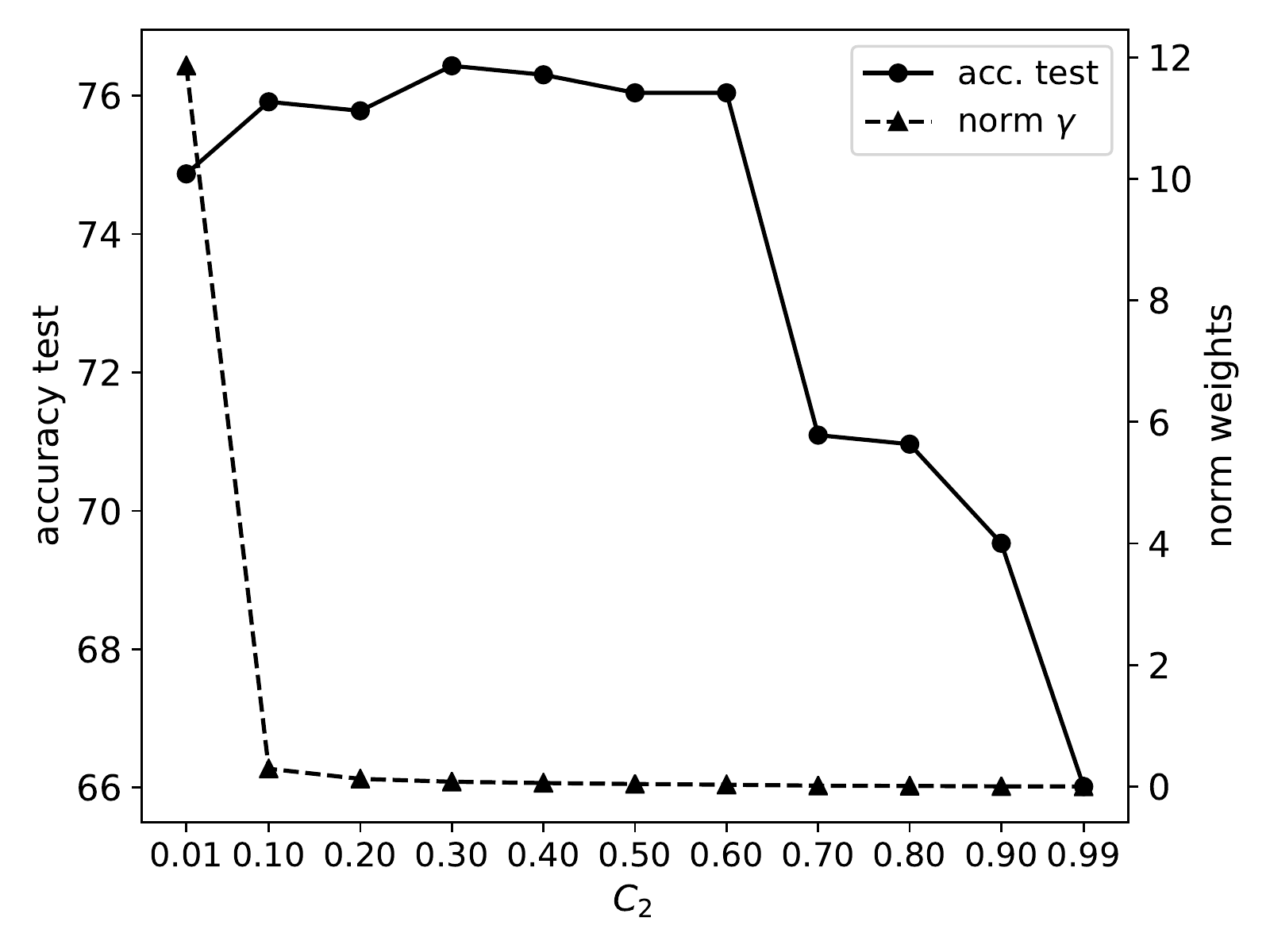}
}
  \hspace{0mm}
  \subfloat[lymphoma]{
  \includegraphics[scale = 0.3]{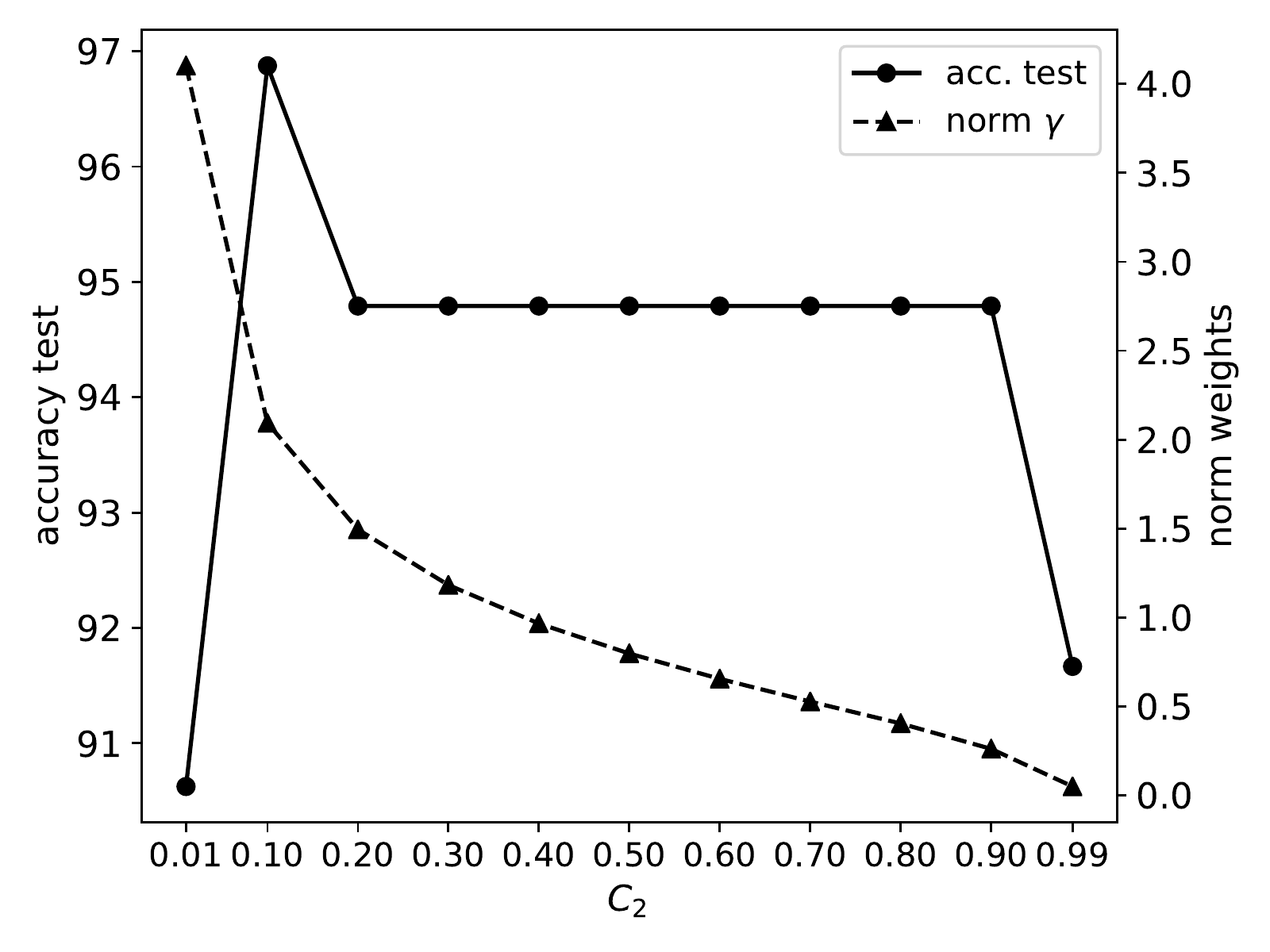}
}
\subfloat[colorectal]{
  \includegraphics[scale = 0.3]{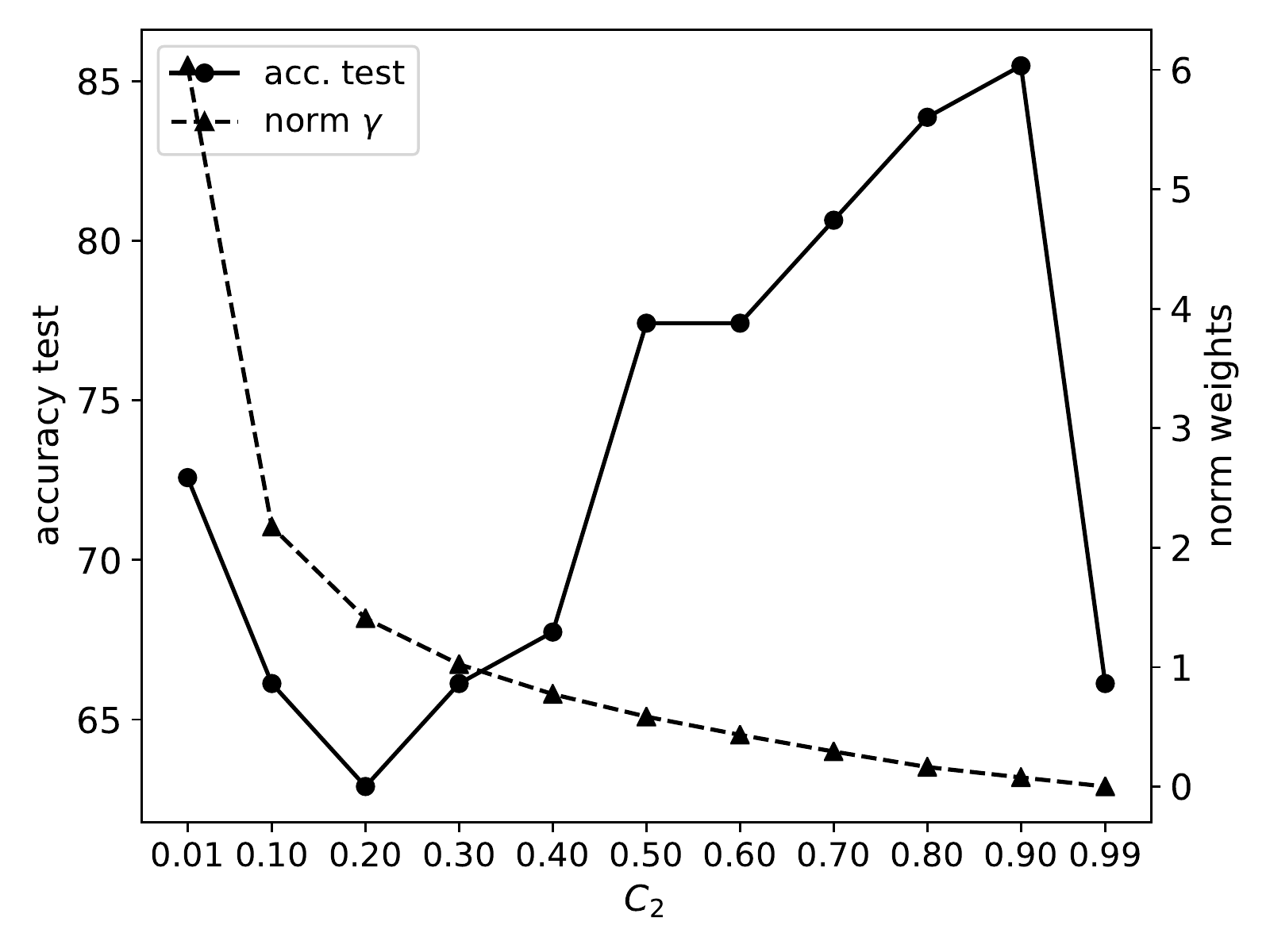}
  \label{subfig:  accuracy test vs norm weights colorectal}
  }
\caption{Plot of the percentage of well-classified on sample $\mathcal{S}_{test}$ over $10$ folds versus the $1$-norm of weights $\gamma$.}
\label{fig: accuracy test vs norm weights}
\end{figure}

  As it was stated in Section \ref{subs: Description of the Experiments}, we do not provide the output results for a single value of $C_2$, but for all the values of $C_2$ in a prefixed grid. In particular, Figure \ref{fig: accuracy test vs norm weights} shows a curve for all the databases where the norm of the weights $\gamma$ versus the percentage of well classified over the $10$ folds is represented\revisionone{, for $C$ in the grid $\{10^{-4}, \ldots, 10^{-1}, 1, 2, \ldots, 9, 10, \ldots, 10^{4}\}$}. Unsurprisingly, in the four data sets, we can see that the larger the value of $C_2$, the lower the norm of the weights $\gamma$. 
  On the other hand, the averaged test accuracy does not feature such a smooth behavior in terms of the $C_2$ values as the $\gamma$-norm does. This is due to two reasons. First, because the second term of the objective function \eqref{subeq: objective function sle} does not maximize the accuracy,  but a proxy of it given by the SVM  margin. Second, the performance of model \eqref{eq: single level equivalent gradient equal zero} is optimized over the sample data $\tilde{\mathcal{S}}$, which provide an incomplete view of reality. Consequently, this performance may not necessarily generalize to the test set, $\mathcal{S}_{test}$. \revisionone{
  This erratic behavior is especially noticeable in the \emph{colorectal} data set (Figure \ref{subfig:  accuracy test vs norm weights colorectal}), which is known to contain outliers, \cite{bolon2014review}. The experiments performed in this paper show that the presence of such atypical data in the test sample $\mathcal{S}_{test}$ distorts the accuracy of our method compared with a test sample where no outliers appear.}
  
  The main take-away of these figures is that values of $C_2$ very close to 0 or 1 lead to low accuracy levels in the test set. Low values of $C_2$ yield complex models that tend to overfit the data and reduce out-of-sample performance. This is especially noticeable in data sets \emph{colorectal} and \emph{lymphoma}, where the ratio between the number of features and the number of individuals is larger. Conversely, high values of $C_2$ produce oversimplified models unable to capture all the explanatory power of the available features.
  
Table \ref{table: accuracy results} shows the best accuracy results in the test sample with our approach together with the $C_2$ value at which such accuracy is reached. We also include the accuracy results obtained in the test samples for all the comparative algorithms presented in Section \ref{subs: Comparative Algorithms}. We recall that the sample division used in methodologies \emph{NO-FS}, \revisionone {\emph{$\ell_1$-SVM},} \emph{KP-FS off-the-shelf} and \emph{MILP-FS off-the-shelf} is exactly the same as that considered in our \emph{MM-FS} approach. Moreover, the accuracy of the methods \emph{KP-FS ad-hoc} and \emph{MILP-FS ad-hoc} has been directly taken from \cite{maldonado2011simultaneous} and \cite{labbe2019mixed}, respectively. In this vein, the \emph{lymphoma} data set has not been used in \emph{MILP-FS ad-hoc}, and therefore, its accuracy result is not available. In addition, databases \emph{colorectal} and \emph{lymphoma} cannot be solved by the \emph{KP-FS off-the-shelf} approach within a time limit of $24$ hours.

  \begin{table}[!htb]
%\hspace*{-1.5cm}
 \centering
% \begin{small}
 \begin{tabular}{ccccc}
 \toprule
  & breast&  diabetes & lymphoma & colorectal\\
 \midrule
\text{MM-FS}& 97.35\% &76.43\% &96.87\% & 85.48\%\\
$C_2$ accuracy MM& 0.7 & 0.3&0.1&0.9\\
\text{NO-FS}& 97.89\%   & 77.08\%& 94.79\% & 83.87\% \\
\revisionone{$\ell_1$\text{-SVM}}& \revisionone{96.83\%  } & \revisionone{76.95\%}& \revisionone{95.83\%} & \revisionone{75.80\%} \\
\text{KP-FS ad-hoc}& 97.55\%  &76.74\% &99.73\% & 96.57\%\\
\text{KP-FS off-the-shelf}& 62.74\% &65.10\% &Max time &Max time\\
\text{MILP-FS ad-hoc}&97.72\%  & 77.75\% & Not avail.& 92.08\%\\
\text{MILP-FS off-the-shelf}&96.48\%  &67.44\% &97.91\%& 85.48\%\\
 \bottomrule
 \end{tabular}
 \caption{Accuracy results for our MM-FS approach and all the comparative algorithms.}\label{table: accuracy results}
% \end{small}
\end{table}
We observe that the proposed \emph{MM-FS} approach obtains similar accuracy results than those achieved with the \emph{NO-FS} method, where all the features have the same weights. In other words, our approach is able to successfully extract the relevant information of the data by selecting the features with the highest classification power. Interestingly, in some cases, such as for the databases \emph{colorectal} and \emph{lymphoma}, the elimination of irrelevant features by our approach improves the prediction accuracy.

\revisionone{Furthermore,  the proposed \emph{MM-FS} procedure performs similarly to the well-known $\ell_1$-SVM in all databases, except for \emph{colorectal}, for which the accuracy of our approach is $10\%$ higher than that of $\ell_1$-SVM. More important, though, is the fact that, even in those databases for which the accuracy of both methods is comparable, the number of features retained by \emph{MM-FS} is notoriously smaller than that of $\ell_1$-SVM. We further elaborate on this later in Section \ref{subs: Feature selection results}.}

Remarkably, our approach, which makes use of available off-the-shelf optimization software, delivers results that are comparable to those given by the ad-hoc implementations of \emph{KP-FS} and \emph{MILP-FS} for all databases, except for \emph{colorectal}, which will be later discussed.

With respect to the comparative algorithm \emph{KP-FS off-the-shelf}, the results reveal that our \emph{MM-FS} approach is significantly better in databases \emph{breast} and \emph{diabetes}. In particular, when the \emph{KP-FS} model is solved with off-the-shelf software, it turns out that the $\gamma$ variables of all features tend to zero, leading to inaccurate predictions, where all the elements are classified with the label of the predominant class. In the case of \emph{lymphoma}, it is even impossible to obtain a local optimal solution within the time limit of $24$ hours. The results obtained with the \emph{MILP-FS off-the-shelf} are slightly better than those of the \emph{MM-FS} methodology for the \emph{lymphoma} data set. In contrast, our proposal is slightly better than \emph{MILP-FS off-the-shelf} for the \emph{breast}. Finally, it can be seen that our \emph{MM-FS} approach delivers significantly better results than \emph{MILP-FS off-the-shelf} for the \emph{diabetes} data set. In fact, the \emph{MILP-FS off-the-shelf} just predicts correctly the predominant class. %This is most likely due to the fact that \emph{MILP-FS off-the-shelf} is limited to \emph{linear} classifiers, while the \emph{diabetes} data set is, however, not linearly separable, as indicated in Section \ref{subs: Data Sets Description}.

We conjecture that the differences observed for data set \emph{colorectal} when comparing our \emph{MM-FS} strategy with algorithms \emph{KP-FS ad-hoc} and \emph{MILP-FS ad-hoc} are due to the significant number of outliers that this data set, of only 62 individuals, contains, \cite{bolon2014review}. Unfortunately, there is no comment on the treatment of outliers in \cite{maldonado2011simultaneous} or \cite{labbe2019mixed}. However, our conjecture is based on the following two facts: i) if these outliers are removed from the data set, the performance of our approach is comparable to that reported for \emph{KP-FS ad-hoc} and \emph{MILP-FS ad-hoc} in \cite{maldonado2011simultaneous} and \cite{labbe2019mixed}, respectively; and ii) the estimated accuracy of the \emph{off-the-shelf} variant of \emph{MILP-FS} is the same as that of our method.

Finally, we do not report an estimated accuracy value for method \emph{KP-FS off-the-shelf}, since it provides no solution within the 24-hour limit.

  \subsection{Feature selection results}\label{subs: Feature selection results}
  In this section, we assess the ability of our approach to select the most relevant features in a data set without impairing, as much as possible, the classification accuracy.
  \revisionone{For this purpose, we next show results from a series of experiments in which we analyze the out-of-sample performance of the nonlinear SVM problem \eqref{eq: dual SVM} that considers only the most relevant features identified by our \emph{MM-FS} method. Following the criterion in \cite{labbe2019mixed}, we consider that a feature is relevant if the associated $\gamma_j$ value provided by our method is greater than $10^{-2}$. Accordingly, the features whose value $\gamma_j$ is less than or equal to $10^{-2}$ are discarded.}
  
 \revisionone{For the sake of comparison, we also conduct the same experiment with $\ell_1$-SVM, which is the standard method used for feature selection in SVM. For consistency, we first employ $\ell_1$-SVM to identify those features such that the absolute value of their associated coefficient $w_j$ is greater than $10^{-2}$. We then estimate the out-of-sample accuracy of $\ell_1$-SVM by recomputing the classifier that this method provides with those selected features only.}
 
\begin{figure}[!htb]
    \captionsetup{labelfont={color=\mylabelcolorrevisedfiguresandtables}}
\thisfloatpagestyle{empty}
%\centering
\hspace*{-2cm}
\subfloat[\revisionone{breast}]{
  \includegraphics[scale = 0.5]{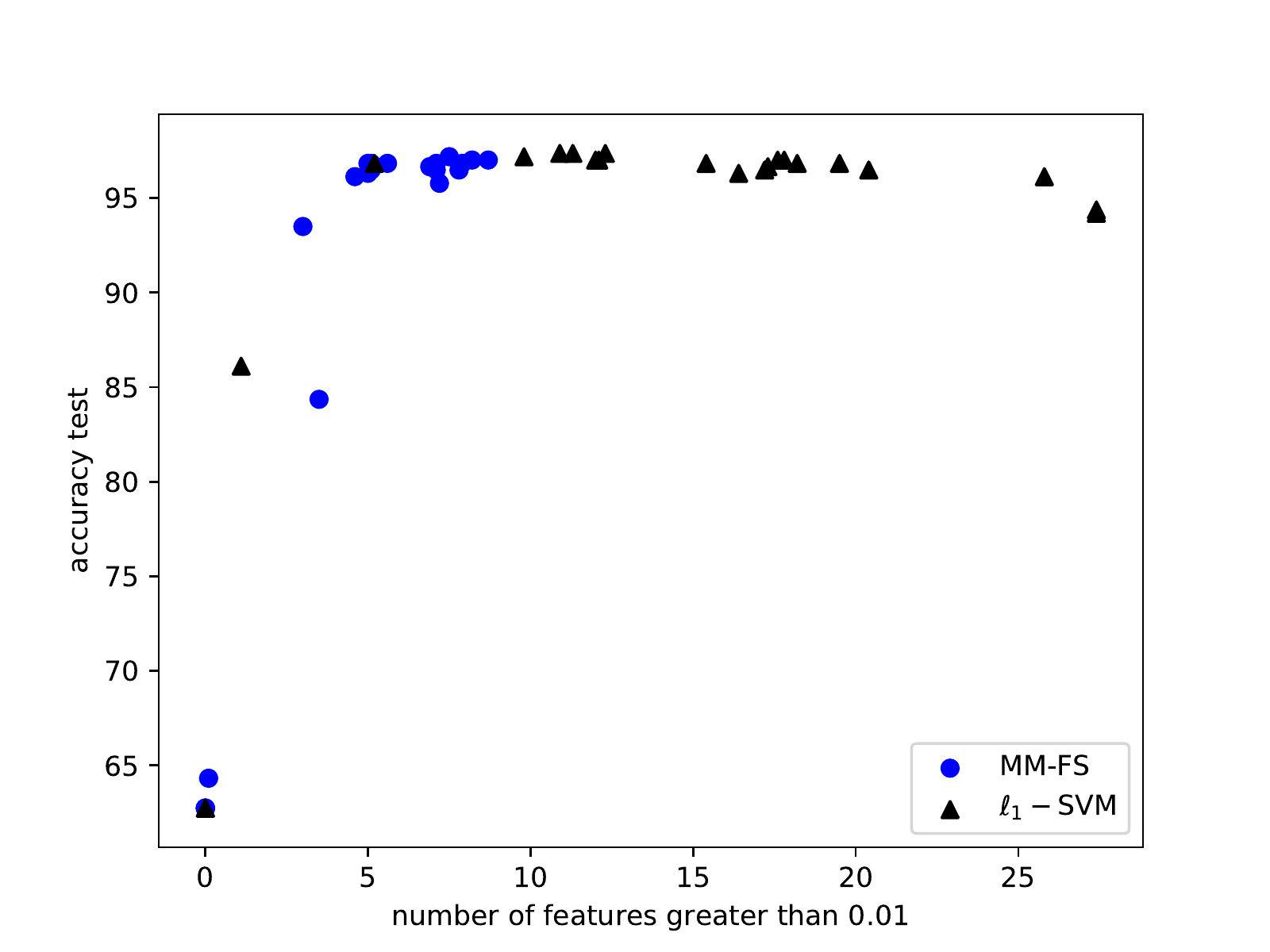}
  \label{}
}
\subfloat[\revisionone{diabetes}]{
  \includegraphics[scale = 0.5]{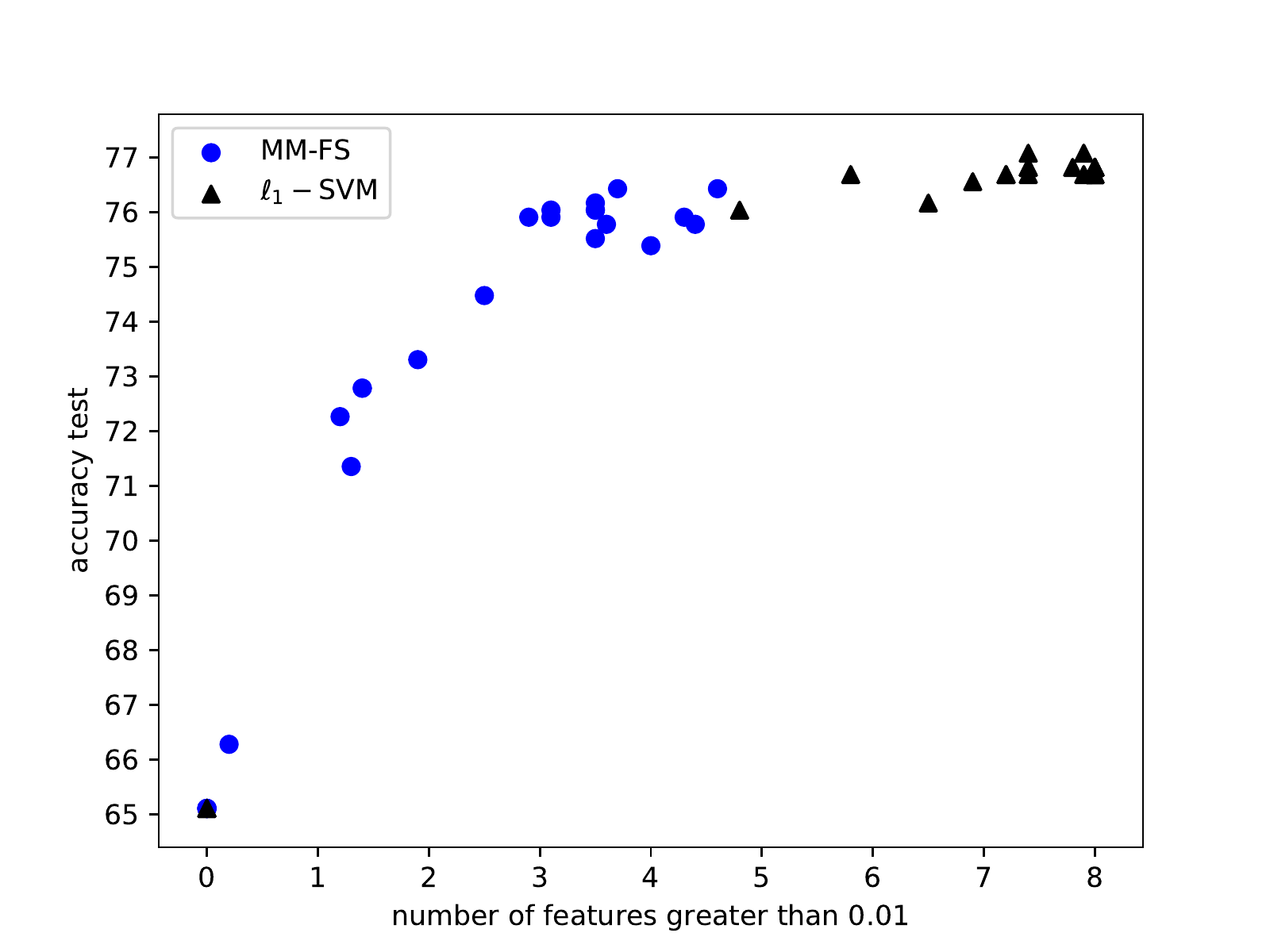}
}
  \hspace{0mm}
  \hspace*{-2cm}
  \subfloat[\revisionone{lymphoma}]{
  \includegraphics[scale = 0.5]{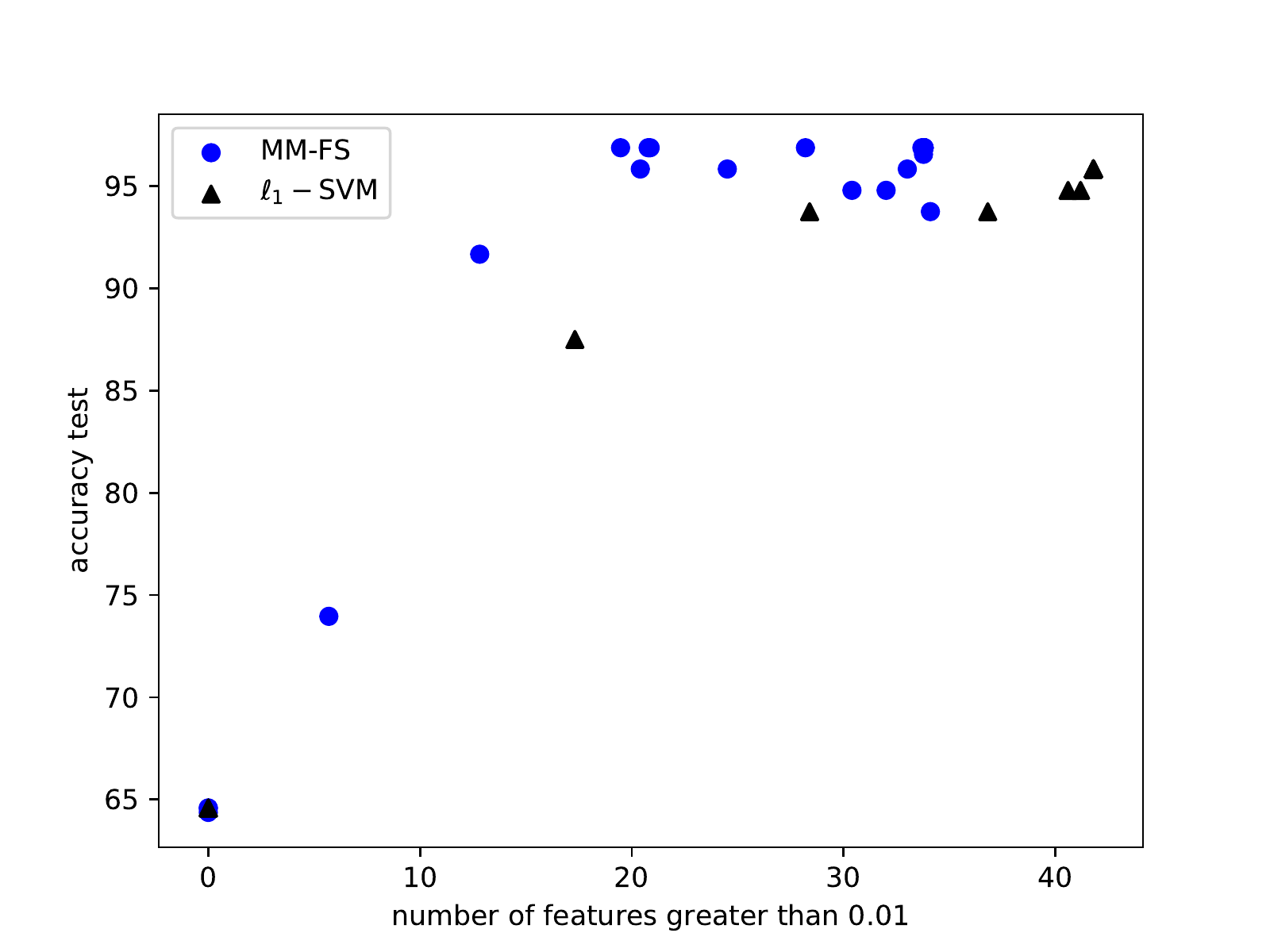}
}
\subfloat[\revisionone{colorectal}]{
  \includegraphics[scale = 0.5]{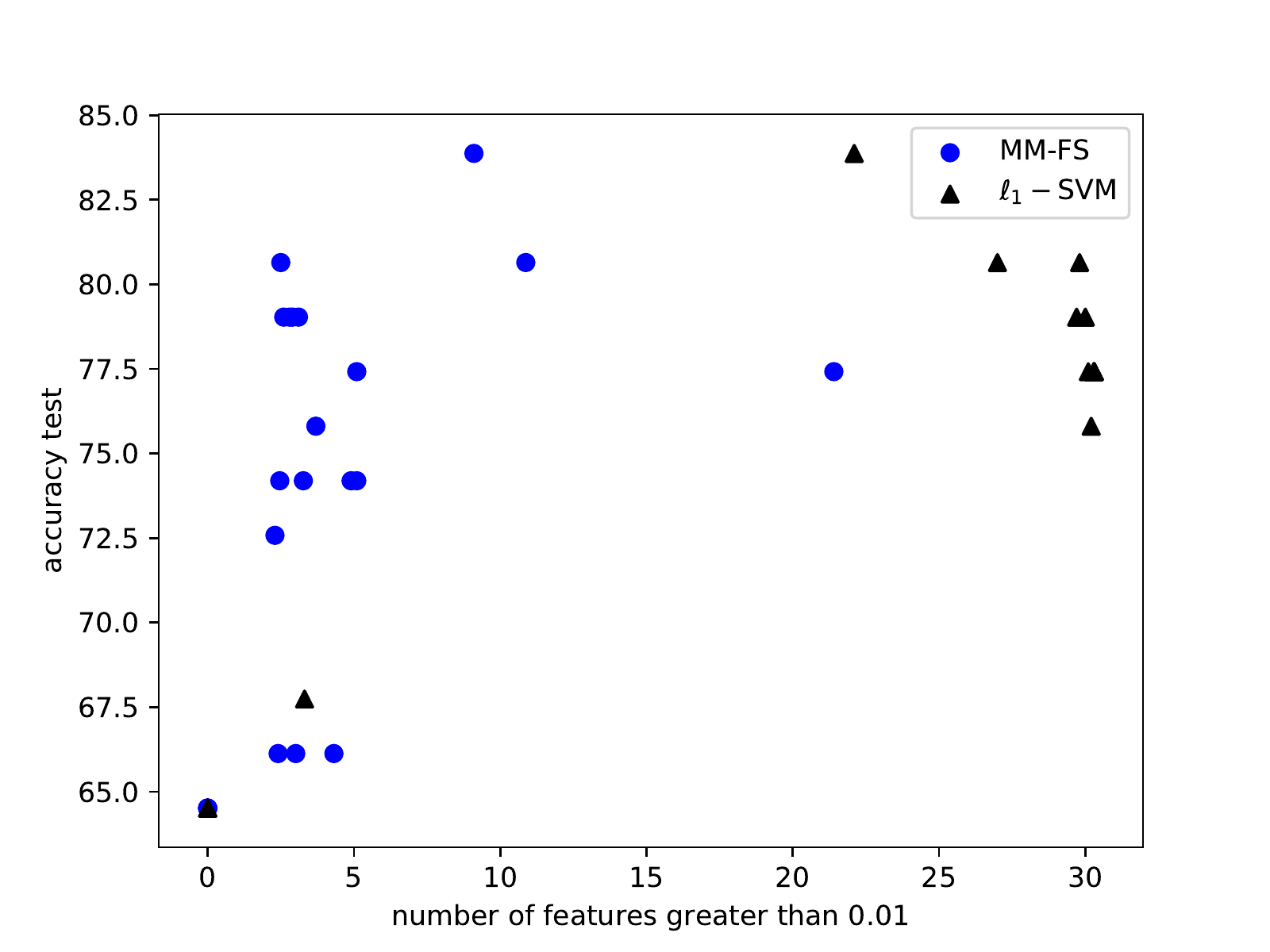}
  }
\caption{\revisionone{Plot of the out-of-sample accuracy versus the number of features selected by \emph{MM-FS} and $\ell_1$-SVM}}
\label{fig: accuracy test vs number of selected features}
\end{figure}

  \revisionone{Figure \ref{fig: accuracy test vs number of selected features} shows the number of features with $\gamma_j$ or $|w_j|$ greater than $10^{-2}$ versus the attained out-of-sample accuracy averaged over all folds. The blue circles and the black triangles refer to \emph{MM-FS} and $\ell_1$-SVM in that order. The optimal vector $\gamma$ provided by \emph{MM-FS} has been obtained for the $C_2$ values given in the second row of Table \ref{table: accuracy results}. Each circle or triangle corresponds to a specific value of the hyperparameter $C$, which has been taken from the discrete set  $\{10^{-4}, \ldots, 10^{-1}, 0.2, \ldots, 0.9, 1, 2, \ldots, 9, 10, \ldots, 10^{4}\}$ to represent a sufficient number of points in the plots. We remark that two different values of $C$ may result in the same point on the plane \# \emph{features greater than 0.01} vs \emph{accuracy test}, thus producing overlapped circles or triangles.}
  
  \revisionone{It can be seen in Figure \ref{fig: accuracy test vs number of selected features} that either \emph{MM-FS} provides a more accurate classifier than that given by $\ell_1$-SVM, or a classifier with a similar level of accuracy but with a significantly lower number of features. For instance, for the database \emph{diabetes}, our approach is able to get an accuracy similar to that of $\ell_1$-SVM, but with approximately half of the features.}

  \revisionone{\subsection{Ranking of features and multicollinearity results}\label{subs: Ranking of features and multicollinearity results}}
  
  \begin{table}[!htb]
  \captionsetup{labelfont={color=\mylabelcolorrevisedfiguresandtables}}
\hspace*{-1cm}
 \centering
 \begin{small}
 {\color{\mylabelcolorrevisedfiguresandtables}
 \begin{tabular}{cccccccccccc}
 \toprule
 $C_2$ & 0.01&  0.1 & 0.2 & 0.3 & 0.4 & 0.5 & 0.6 & 0.7 & 0.8 & 0.9 & 0.99\\
 \midrule
\multirow{5}{*}{breast}& 21 &21&21 &21&21&21&21&21&21&21&28\\
&11&22&22&22&22&22&22&22&22&28&21\\
&22&11&8&25&29&28&25&25&28&22&8\\
&25&7&25&29&25&29&29&29&25&8&23\\
&30&25&11&8&28&25&28&28&8&7&22\\
\midrule
\multirow{5}{*}{diabetes} &2&2&2&2&2&2&2&2&2&2&1\\
&8&6&6&6&6&6&6&1&1&1&2\\
&1&7&1&1&1&1&1&6&6&3&3\\
&6&8&7&7&7&7&3&3&3&6&4\\
&7&1&3&3&3&3&7&7&7&4&5\\
\midrule
\multirow{5}{*}{lymphoma}&   461&3783&3783&3783&3783&3783&3783&3783&3783&3783&3794\\
&2267&461&3794&512&512&512&512&512&512&512&2251\\
&512&2267&512&3794&3794&3794&3794&3794&3794&3794&2493\\
&237&2251&2251&2251&2251&2251&2251&2251&2251&1941&3783\\
&3119&512&461&461&461&236&461&461&461&2251&512\\
\midrule
\multirow{5}{*}{colorectal}& 43&377&377&765&765&377&377&377&377&377&1\\
&377&43&765&1772&1772&1772&286&1772&1993&1870&2\\
&1325&765&1772&377&377&765&1772&765&66&249&3\\
&1993&1772&70&792&70&286&765&286&1641&1772&4\\
&1241&974&792&70&792&341&1641&1993&765&1993&5\\
 \bottomrule
 \end{tabular}}
 \caption{\revisionone{Ranking of the features by the proposed approach \emph{MM-FS} for different values of $C_2$.}}
 \label{table: ranking feature results}
 \end{small}
\end{table}
  
  \revisionone{The first purpose of this section is to study how consistent our method is when ranking features for different values of $C_2$, as explained in Section 4.2. Table \ref{table: ranking feature results} shows the five most important features that are identified by our approach for each database and for different $C_2$ values. For instance, \emph{MM-FS} with $C_2 = 0.5$ in the \emph{breast} data set yields that the first and fifth most relevant features are 21 and 25, respectively. We can observe that, in all data sets, there exists a set of features which is selected to be important no matter if a lower or higher penalization is applied. Hence, we can conclude that the proposed feature selection methodology is robust against the choice of the hyperparameter $C_2$.}

 \begin{figure}[!htb]
 \captionsetup{labelfont={color=\mylabelcolorrevisedfiguresandtables}}
\centering
  \includegraphics[scale = 0.8]{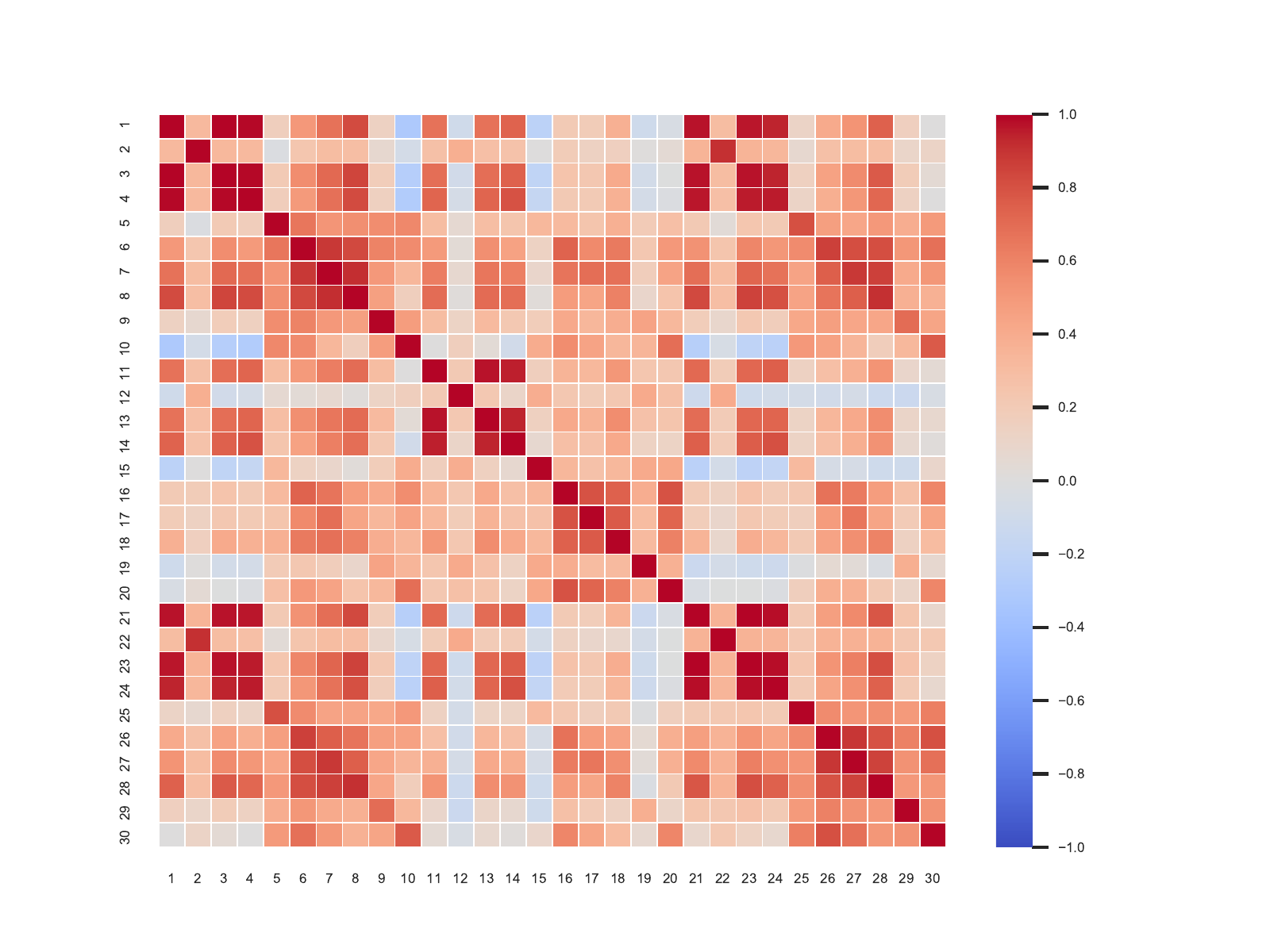}
\caption{\revisionone{Correlation matrix of database \emph{breast}}}
\label{fig: correlation_matrix_breast}
\end{figure}

\revisionone{We next discuss the effect of multicollinearity when selecting features by our approach. Figure \ref{fig: correlation_matrix_breast} depicts the correlation matrix of the thirty features involved in the \emph{breast} data set. If we pay attention, for instance, to the top-five features selected by our approach with $C_2 = 0.7$ (Table \ref{table: ranking feature results}), we can observe that their associated correlation values are relatively low. Furthermore, Figure  \ref{fig: correlation_matrix_breast} shows that the features belonging to the set $\{1, 3, 4, 21, 23, 24\}$ are highly correlated. However, only feature 21 from this set has been selected as important by our approach. Hence, it is proven that the proposed \emph{MM-FS} method just selects a single feature among a set of correlated ones instead of selecting the whole set of important and correlated features. Similar conclusions can be drawn from the rest of the databases.}
\\

   \subsection{Interpretability of the results}\label{subs: Interpretability of the results}
  
   To evaluate the interpretability of \emph{MM-FS}, we compare in Table \ref{table: selected features benchmark} the five most important features that our approach identifies with respect to the relevant features pinpointed in the technical literature. \revisionone{For each database, the results associated with our approach correspond to the $C_2$ value that delivers the highest classification accuracy, as Table \ref{table: accuracy results} shows. For the sake of interpretability, the features of databases \emph{breast}, \emph{diabetes} and \emph{colorectal} are denoted using their name as well as their identification number, written between parentheses. This number coincides with that given in Table \ref{table: ranking feature results}. Moreover, the features} are presented according to their relevance. For instance, for the \emph{breast} data set, features \emph{radius\_worst} and \emph{texture\_worst} are the most and the second most important features. We have included a reference between parentheses to the benchmark approach used for all databases, except for \emph{lymphoma} since no article reports the most relevant features for this database. 
   %To the best of our knowledge, there is no article which reports the most relevant features in the \emph{lymphoma} data set or the name of their features. Hence, in this case, we just include the identification number of the selected features.
  
    \begin{table}[!htb]
\hspace*{-1cm}
 \centering
  \begin{small}
 \begin{tabular}{c|cccc}
 \toprule
  & breast & diabetes & lymphoma &  colorectal \\
 \hline
\multirow{5}{*}{\text{MM-FS}}&radius\_worst \revisionone{(21)},    &glucose \revisionone{(2)}, &3783, & Hsa.36689 \revisionone{(377)},  \\
& texture\_worst \revisionone{(22)},&body\_mass\_index \revisionone{(6)},& 461,  & Hsa.1660 \revisionone{(1870)},\\
&smoothness\_worst \revisionone{(25)}, &pregnancies \revisionone{(1)},&2267,& Hsa.8147 \revisionone{(249)}, \\
&symmetry\_worst \revisionone{(29)}, & diabetes\_pedigree \revisionone{(7)},&2251,&Hsa.6814 \revisionone{(1772)},\\
&concave\_points\_worst \revisionone{(28)}& blood\_pressure \revisionone{(3)}&512&Hsa.41260 \revisionone{(1993)}\\
 \hline
\multirow{4}{*}{\text{Bench}}&radius\_worst \revisionone{(21)},   &glucose \revisionone{(2)}, &\multirow{4}{*}{ Not avail.}& Hsa.36689 \revisionone{(377)},\\
& texture\_worst \revisionone{(22)}, & body\_mass\_index \revisionone{(6)}, &&Hsa.37937 \revisionone{(493)},\\
&concave\_points\_worst \revisionone{(28)} &diabetes\_pedigree \revisionone{(7)}&&Hsa.6814 \revisionone{(1772)}\\
&(Ref. \cite{ghazavi2008medical})& (Ref. \cite{yang2018deep})&&(Ref. \cite{li2002bayesian})\\
 \bottomrule
 \end{tabular}
 \caption{Comparison of the selected features in the \emph{MM-FS} approach and some benchmark approaches from the technical literature.}\label{table: selected features benchmark}
\end{small}
\end{table}
 
 We can observe in Table \ref{table: selected features benchmark} that most of the features that our methodology identifies as relevant are also deemed as significant in the technical literature. In the case of the \emph{breast} data set, besides identifying already known important features, \cite{ghazavi2008medical}, our approach also considers as relevant features \emph{smoothness\_worst} and \emph{symmetry\_worst}. Something similar happens in the data set \emph{diabetes}, where features \emph{glucose}, \emph{body\_mass\_index} and \emph{diabetes\_pedigree} among others are selected to be relevant for determining whether a new patient suffers from diabetes or not. This conclusion coincides with the one obtained in the literature, \cite{yang2018deep}. Finally, the model here proposed is able to found $2$ of the genes which have been detected in the literature, \cite{li2002bayesian}, to be important in the diagnosis of colon cancer out of the $2000$ features available in the data set \emph{colorectal}. Apart from this information, the \emph{MM-FS} methodology also selects three more relevant genes.
 
  Hence, our proposal is competitive compared to benchmark methodologies, not only in terms of classification accuracy and feature ranking, as Sections \ref{subs: Accuracy results} and \ref{subs: Feature selection results} respectively show, but also in terms of interpretability.

  \section{Conclusions and Future Work}\label{sec: Conclusions and Future Work}
  This paper deals with the problem of feature selection in nonlinear SVM classification. To this aim, a novel embedded feature selection approach is proposed, by means of a min-max optimization problem that seamlessly balances model complexity and classification accuracy. Unlike existing ad-hoc approaches, the proposed model can be efficiently solved with standard off-the-shelf optimization software, thanks to an equivalent reformulation that leverages duality theory. 
  
 Numerical experience shows that our feature selection approach is able to select and rank the features in terms of their predictive power, preserving similar out-of-sample accuracy results than the classification performance obtained when all the features are considered. Besides, numerical tests with various databases show that the proposed approach significantly outperforms state-of-the-art embedded methods for feature selection when solved with off-the-shelf software and is comparable to them when these are solved using ad-hoc strategies. \revisionone{We checked that the proposed strategy is consistent with respect to the ranking of features that it provides for different $C_2$ values, and studied the effect of multicollinearity on our model.} Finally, our approach produces interpretable classification models and correctly identify the relevant features reported in the literature.  
 
In this paper, we have restricted ourselves to the Gaussian kernel. Nevertheless, the model here proposed can be extended to other families of kernels, such as polynomial or sigmoid in a  straightforward manner.  In addition, the extension of our proposal to other Data Science problems, e.g., regression or clustering, or to other real-world applications, for instance, for power system operations, deserves further study.
\revisionone{Besides, the application of evolutionary algorithms in the line of \cite{xue2016survey,xue2019self, zhang2017multi} to tackle hard-to-solve SVM models for feature selection could be further explored too.}

\section*{Acknowledgments}
This work was supported in part by the Spanish Ministry of Economy, Industry, and Competitiveness through project ENE2017-83775-P, in part by the European Research Council (ERC) under the EU Horizon 2020 research and innovation program (grant agreement No. 755705) and in part, by the Junta de Andalucía (JA), the Universidad de Málaga (UMA), and the European Regional Development Fund (FEDER) through the research project UMA2018‐FEDERJA‐001. The authors thankfully acknowledge the computer resources, technical expertise and assistance provided by the SCBI (Supercomputing and Bioinformatics) center of the University of Málaga.
  
\bibliography{\myreferences}

\end{document}